\documentclass[default]{sn-jnl}

\usepackage{amsmath}
\usepackage{flexisym}
\usepackage{dsfont}
\usepackage{tabularx}
\usepackage{longtable,lscape}
\usepackage[ruled]{algorithm2e}
\usepackage{algpseudocode}
\usepackage{natbib}

\jyear{2021}%

\theoremstyle{thmstyleone}%
\theoremstyle{thmstyletwo}%

\theoremstyle{thmstylethree}%

\begin{document}

\title[Neural Networks for Scalar Input and Functional Output]{Neural Networks for Scalar Input and Functional Output}


\author[1]{\fnm{Sidi} \sur{Wu}}\email{sidi\_wu@sfu.ca}

\author[2]{\fnm{C\'edric} \sur{Beaulac}}\email{beaulac.cedric@uqam.ca}

\author*[1]{\fnm{Jiguo} \sur{Cao}}\email{jiguo\_cao@sfu.ca}

\affil[1]{\orgdiv{Department of Statistics and Actuarial Science}, \orgname{Simon Fraser University}, \orgaddress{\city{Burnaby}, \state{BC}, \country{Canada}}}

\affil[2]{\orgdiv{Département de Mathématiques}, \orgname{Université du Québec à Montréal}, \orgaddress{\city{Montréal}, \state{Québec}, \country{Canada}}}


\abstract{The regression of a functional response on a set of scalar predictors can be a challenging task, especially if there is a large number of predictors, or the relationship between those predictors and the response is nonlinear. In this work, we propose a solution to this problem: a feed-forward neural network (NN) designed to predict a functional response using scalar inputs. First, we transform the functional response to a finite-dimensional representation and construct an NN that outputs this representation. Then, we propose to modify the output of an NN via the objective function and introduce different objective functions for network training. 
The proposed models are suited for both regularly and irregularly spaced data, and a roughness penalty can be further applied to control the smoothness of the predicted curve.
The difficulty in implementing both those features lies in the definition of objective functions that can be back-propagated. In our experiments, we demonstrate that our models outperform the conventional function-on-scalar regression model in multiple scenarios while computationally scaling better with the dimension of the predictors.}

\keywords{Functional data analysis, Functional response, Functional principal component analysis, Machine learning}



\maketitle

\section{Introduction}\label{sec:intro}

\begin{figure}[ht]
    \centering
    \includegraphics[width = 0.95\columnwidth]{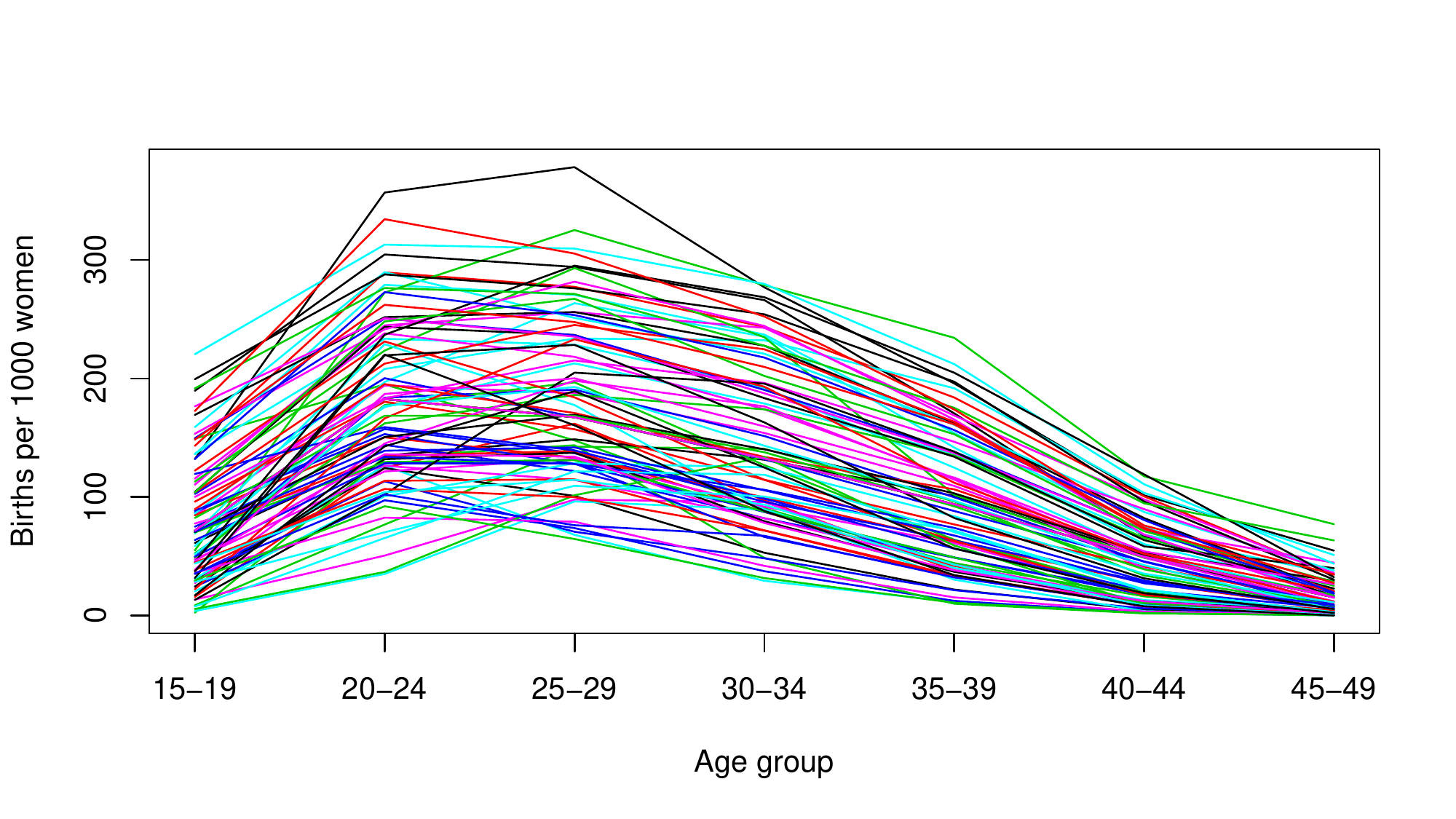}
    \caption{Age-specific fertility rates for 92 countries around the world.}
\label{fig:fertility_curves}
\end{figure}

Functional data analysis (FDA) is a rapidly developing branch of statistics which targets at the theory and analysis of functional variables. As the atom of FDA, functional variables or functional data refer to curves, surfaces and any random variables taking values in an infinite dimensional space, such as time and spatial space \citep{fda, nonparametric_fda}. A basic and commonly recognized framework in FDA is to regard the functional data as realizations of an underlying stochastic process \citep{fda_review}, and indeed, a large fraction of data coming from different fields can be characterized as functional data.  Figure \ref{fig:fertility_curves}, for example, illustrates the fertility rates over the age of females for 92 countries. Each curve is treated as a smooth function of age and can be viewed as functional data. The fertility curves were measured at several time points, each of which represents an age group in increasing order. This age-specific fertility rate (ASFR) data was collected from the United Nations Gender Information (UNGEN) database by Mehrotra and Maity \citep{real_data}. They estimated the observations based on the surveys conducted between 2000 and 2005. The data set is publicly available at \url{https://github.com/suchitm/fosr_clust}.

Among different aspects of FDA, functional regression has received the most attention in applications and methodological development \citep{functionalregression}. Functional regression, in general, can be divided into three types: (1) scalar-on-function regression (regression analysis of scalar responses on functional predictors)\citep{Lin17}, (2) function-on-function regression (predictors and responses of the regression are both functional) \citep{cai2021,cai2021stat}, and (3) function-on-scalar (FoS) regression (regression of functional responses on scalar predictors, also named as the varying-coefficient model)\citep{cai2022}. While plenty of work has been done for the first two scenarios, studies discussing functional response regression remain scarce. Functional linear models (FLM), a series of extensions of the classic linear models, are the most conventional and widely applied methods in dealing with regression problems. A general form of FLM for FoS regression can be written as 
\begin{align}
    Y(t) = \boldsymbol{X}\boldsymbol{\beta}(t)+\epsilon(t),
\end{align}
where $Y(t)$ is the functional response, and $\boldsymbol{X}$ is a vector of scalar covariates. The unknown vector $\boldsymbol{\beta}(t)$ consists of parameter functions varying over $t$, and $\epsilon(t)$ stands for the random error term. In FDA, the consensual approach is to represent functional data with a linear combination of a finite number of known basis functions. Ramsay \& Silverman \citep{fda} introduced projecting functional response with basis functions like B-splines for fitting the FoS regression model. Chiou et al. \citep{Chiou2004, Chiou2003} proposed to use functional principal component analysis (FPCA) for dimensional reduction and represent the functional response with eigenfunctions obtained via spectral decomposition of the covariance function of $Y(t)$. Both approaches summarize the information carried by the functional variable $Y(t)$ to a finite-dimensional vector of scalar representations (B-spline basis coefficients or functional principal component scores (FPC scores)). In solving linear functional regression problems, taking FoS regression as an example, the linear relation between the scalar predictors and the functional response is indeed captured by fitting a linear relation between the predictors and the scalar representation. 

Some nonlinear approaches have been developed to handle more complicated regression settings. Zhang and Wang \citep{VCAM_Zhang} combined the FoS with the additive models to gain the varying-coefficient additive model for functional data which does not require the linearity assumption between the scalar predictors and the functional response. Afterwards, an extension to this varying-coefficient additive model, named as functional additive mixed (FAM) model, was established by Scheipl et al. \citep{FAMM_Scheipl} with a general form
\begin{align}
    Y(t) = \sum_{r=1}^{R}f_{r}\left(\boldsymbol{X}_{r}, t\right) + \epsilon(t),
\end{align}
for functional response $Y(t)$. Each term in the additive predictor $f_{r}(\boldsymbol{X}_{r}, t)$ is a function of $t$ and a subset $\boldsymbol{X}_{r}$ of the complete predictor set $\boldsymbol{X}$ including scalar and functional covariates. This extensive framework can consist of both linear and nonlinear effects of functional and scalar covariates that may vary smoothly over the index of the functional response. Naturally, this model was further extended, by Scheipl et al \citep{GFAMM_Scheipl}, to the generalized functional additive mixed model in order to take care of non-Gaussian functional response. Although several nonlinear attempts have been made, the existing regression methods for FDA have been predominantly linear \citep{fda_review}. Considering the emerging trend of complicatedly structured functional data, there is an increasing demand to develop more nonlinear approaches to FDA.


In this work, we propose a solution to the FoS problem which is able to handle a large number of predictors and the nonlinear relation between the scalar predictors and the functional response. Our solution borrows from the machine learning literature, which follows a trend in adapting machine learning techniques to known statistical problems such as survival analysis \citep{ishwaran2008random, katzman2018deepsurv, beaulac2018deep} or causal inference \citep{scholkopf2021toward, lecca2021machine}. We adapt the neural network (NN) architecture for functional data. 

In some existing works, efforts have been made to combine deep NNs to the field of functional data analysis. For instance, Rossi et al.\citep{Rossi2002, Rossi2005} firstly explored the idea of applying NNs to functional data by constructing a functional neural network (FNN) with functional neurons in the first hidden layer for functional inputs. FNN was then extended by Thind et al. \citep{thind2020deep,thind2020neural} to feed both functional and scalar coviarates as inputs and outputs a scalar response. Rao et al. \citep{Spatio-TemporalFNN} equipped FNN with geographically weighted regression and spatial autoregressive technique to handle regression problems with spatially correlated functional data. Meanwhile, Wang et al.\citep{FoFNN} proposed a nonlinear function-on-function model using a fully connected NN. Previously, Yao et al. \citep{AdaFNN} developed an NN with a new basis layer whose hidden neurons are micro NNs, to perform a parsimonious dimension reduction for functional inputs using information relevant to the scalar target. \citep{WangCao2023} introduced a functional nonlinear learning approach to adequately represent multivariate functional data within a reduced-dimensional feature space. \citep{WangCao23Environmetrics} proposed a nonlinear prediction method for functional time series. 

Most of the mentioned works are focused on building NNs with functional inputs and scalar outputs. In this work, we consider the other side of the coin. We design an NN meant to predict a functional response and to the best of our knowledge, this study is the first attempt to solve the FoS regression problem using artificial NNs. Because the standard machine learning techniques are designed for finite-dimensional feature vectors, we propose to encode the information contained in the intrinsically infinite-dimensional functional response to a finite-dimensional vector of scalar representations. Different from other nonlinear approaches targeting at FoS regression problems, our methods, in particular, focus on studying the nonlinear association between the predictors and the scalar representations of the functional response, to further reveal the relation between the scalar predictors and the functional response. In this way, we maintain the interpretability of the relation between the scalar predictors and the functional response, despite of the usage of NNs.

Challenged by the ASFR data introduced previously, in this work, we are interested in conducting a FoS analysis to accurately predict the functional curves using scalar predictors and also reveal any potential nonlinear relation between the predictors and the response. For each of the 92 observations (countries), the functional curve of fertility rates is associated with 15 demographic and socioeconomic variables averaging over the information available on Gapminder in a country-level manner from 2000 to 2005 \citep{real_data}. These 15 scalar covariates consist of age at first marriage, under-5 mortality, maternity deaths per 1000 women, cervical cancer deaths per 100,000 women, female labor force participation, male to female ratio (women aged from 15-49), contraception prevalence (women aged from 15-49), life expectancy, mean years of school (women \% men, women aged from 15-34), female's body mass index (BMI), the number of births attended by trained birth staff (\% of total), gross domestic product (GDP) per capita, the proportion of dollar billionaires per 1 million people, health expenditures (\% of GDP) and the amount of alcohol its populace consumes. All the covariates downloaded from the listed data source have been previously standardized, and many predictor pairs among them are found highly correlated.

Due to the difficulty in directly determining whether there is a nonlinear relation between a scalar variable and a random function, we pay more attention to the association between the scalar covariates and the scalar representations summarizing the information of the functional response. Regarding the scalar representations, we choose to use the basis coefficients which are estimated by approximating the response function with a linear expansion of the most common B-spline basis functions. Besides the initial choice of the basis coefficients, we also attempt to implant the FPC scores as the scalar representations in the real application.

\begin{figure}[ht]
    \centering
    \includegraphics[width = 0.9\columnwidth]{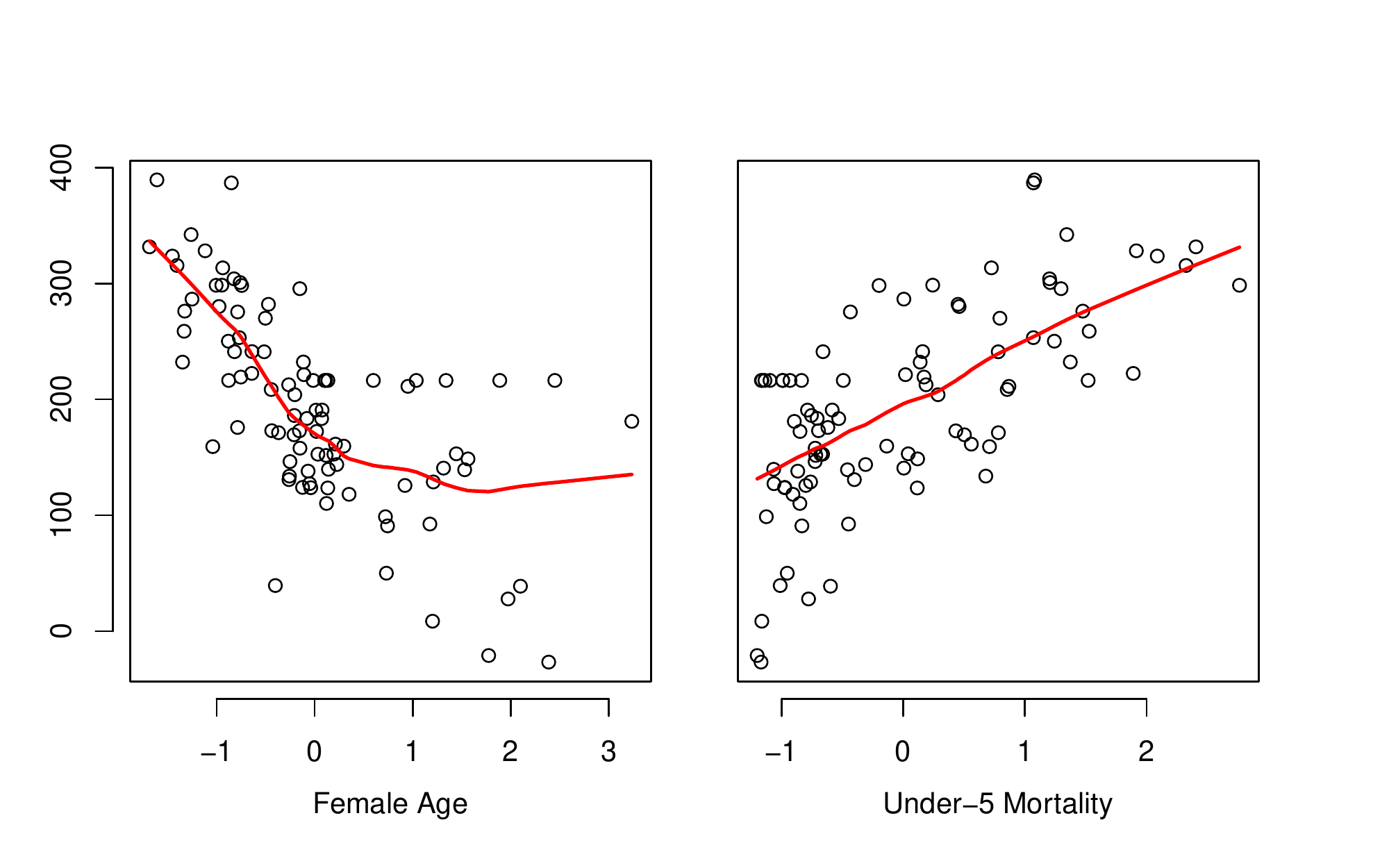}
    \caption{Plots of the second B-spline basis coefficient (estimated using the functional response $Y(t)$) versus two scalar covariates, including female age and under-5 mortality, separately. The red lines are the smoothing curves estimated using \texttt{R} function \texttt{lowess()}.}
    \label{fig:raw_relation_plots}
\end{figure} 

Accordingly, with the ASFR data set, instead of picturing the relation between the infinite-dimensional response curve and the 15 scalar covariates, we obtain visualizations of one B-spline basis coefficient against one predictor for all possible combinations. The representation process is purely based on the fertility trajectories with no contribution from the 15 predictors. Six B-spline basis functions are employed, which is recommended by the cross-validation we performed. Figure \ref{fig:raw_relation_plots} displays the in-pair associations for some selected basis coefficient-predictor pairs. We incorporate a locally-weighted smoothing curve in red to help visualizing if the association is linear or not. The existence of nonlinear associations between some coefficient-predictor pairs is endorsed, as shown with the female age predictor in Figure \ref{fig:raw_relation_plots}. On the other hand, the association pattern of the second basis coefficient to under-5 mortality is basically linear as depicted in Figure \ref{fig:raw_relation_plots}. As both linear and nonlinear patterns are detected among all possible basis coefficient-predictor sets, we would like to set up a procedure with the ability to include every available covariate to precisely predict the response trajectory, while simultaneously capturing the true relationships between the predictors and the scalar representations of the functional response.

Three main contributions are made in this paper. First, we construct a general framework of NN for functional response which aims at predicting curves using the input through a feed-forward NN architecture by slightly modifying the output of the NN via the objective function. Such modifications are readily applicable to other types of networks such as convolutional neural networks (CNN) \citep{zhu2016deep}, and long short-term memory (LSTM) model \citep{LSTM1, LSTM2, LSTM3}. Second, we pay attention to the behavior of the functional output during the network training process and propose to hold their natural smoothness property by adding a roughness penalty in the objective function for training NN. Lastly, we make numerous comparisons under various scenarios to analyze and conclude the conditions of use for different versions of our methods.

The biggest challenge to the implementation of the aforementioned contributions is to ensure the NN models can be trained using backpropagation as a standard NN \citep{rumelhart1986}. Theoretically, we need to define the flow of information from the input to the objective function as a sequence of operations differentiable with respect to the weights of the NN. However, in practice, because we rely on an auto-differentiation package, such as \textbf{Keras} \citep{chollet2015}, to train the NN, we need to define the objective function as a sequence of operations provided by that package. The available differential operations are incapable of dealing with derivatives of the integral of the infinite-dimensional functional response with respect to  NN weights. Besides, some common features when fitting functional data, such as roughness penalties, were obviously not designed with that consideration in mind. For example, it is common to penalize the integrated second-order derivative of a functional curve for smoothing in FDA. But a roughness penalty cannot be simply integrated into the objective function when training the NN. Overcoming that challenge is a central part of these contributions mentioned above and to do so, we needed to be creative and adapt the objective functions in various ways detailed later in this manuscript.

The remainder of this paper is organized as follows. In Section \ref{sec:models}, we detail the proposed NN with a functional response, with an additional discussion on how to control the smoothness of the predicted curves. A brief discussion about the computational costs of NNs and the FLM with functional response is provided in Section \ref{sec:computationalcost}. Section \ref{realapplication} contains the results of prediction in the real application using different methods. In section \ref{simulation}, we conduct simulation studies, under linear and nonlinear scenarios, and compare the proposed models with the existing FoS model in both the predictive accuracy for discretely observed time points and the ability to reconstruct the true response trajectories over a continuum. Lastly, some concluding remarks are given in Section \ref{sec:conclusion}.

\section{Methodology}\label{sec:models}

Suppose we have $N$ subjects and for the $i$-th subject, the input is a set of numerical variables $\boldsymbol{X}_{i} = \{X_{i1}, X_{i2},...,X_{iP}\}$, and the output is a functional variable $Y_{i}(t), t \in \mathcal{T}$ in the $L^{2}(t)$ space. Functional data is often assumed to lie in the $L^2$ space because it ensures that the data is square-intergrable, allowing the optimization of the statistical models with respect to the squared-loss criterion in the FoS regression framework. In reality, the functional response $Y_{i}(t)$ is usually measured in a discrete manner, for instance, at $m_{i}$ time points or locations. Therefore, instead of observing the full trajectory of $Y_{i}(t)$, for the $i$-th subject, we obtain $m_{i}$ pairs of observations $\{t_{ij}, Z_{i}(t_{ij})\}$, $j = 1, 2, ... m_{i}$ and $Z_{i}(t_{ij}) = Y_{i}(t_{ij}) + \epsilon_{i}(t_{ij})$, where $\epsilon_i(t_{ij})$ is the i.i.d. observation error. To simplify the situation, in the following discussion, we assume that for all subjects, the functional term $Y_i(t)$'s are observed at the same $m$ dense and equally-spaced time points. In other words, we assume that $m_{1}=m_{2}=\dots=m_{N}=m$ and $t_{1j}=t_{2j}=\dots=t_{Nj} = t_j$ for $j = 1, 2, ..., m$. A summary of notations used throughout the manuscript is provided in Table S1 in the supplementary document.

\subsection{Neural network with functional response: mapping to basis coefficients (NNBB)}\label{subsec:be}

To overcome the difficulty in modeling infinite-dimensional data, a common pipeline in FDA is to summarize the information of functions $\{Y_i(t)\}_{i=1}^{N}$ into a set of finite-dimensional vectors of coefficients using some basis representation method, and consequently, 
\begin{align}
Y_i(t) = \sum_{k=1}^{K_{b}}c_{ik}\theta_{k}(t)  = \boldsymbol{\theta}' \boldsymbol{C}_{i},
\label{BasisExpansion_EQ}
\end{align}
where $K_{b}$ is the number of basis functions. It is common practice to select $K_b$ by cross-validation \cite{fda}. The vector $\boldsymbol{\theta}$ contains the basis functions $\theta_1(t), ..., \theta_{K_{b}}(t)$ from a selected basis system, such as the Fourier basis system or B-spline system (in our study, we mainly used the latter), and $\theta_{k}(t)$ also belongs to $L^{2}$ space. $\boldsymbol{C_i}$ is  a $K_{b}$-dimensional vector of basis coefficients $\{c_{ik}\}_{k=1}^{K_{b}}$ which lies in $\mathbb{R}^{K_b}$ and needs to be determined. These basis coefficients serve as parameters of this model. Without the smooth underlying functions $Y_i(t)$'s, the discrete observations $Z_i(t_{ij})$'s are used to estimate $\{c_{ik}\}_{k=1}^{K_{b}}$ by fitting $Z_{i}(t_{ij}) = \sum_{k=1}^{K_{b}}c_{ik}\theta_{k}(t_{ij}) + \epsilon_{i}(t_{ij})$. The least square estimator of $c_{ik}$, denoted by ${c}^{\circ}_{ik}$, is obtained by minimizing the sum of squared error (SSE) criterion $\text{SSE}(\boldsymbol{Z}|\boldsymbol{C}) = \sum_{i=1}^{N}\sum_{j=1}^{m}\left(Z_i(t_{ij}) - \sum_{k=1}^{K_{b}}c_{ik}\theta_{k}(t_{ij})\right)^{2}$.

Eq.(\ref{BasisExpansion_EQ}) implies that $Y(t)$ can be approximated by a linear combination of the basis functions $\theta(t)$'s, which carry the fixed modes of variations over $\mathcal{T}$ with real-valued basis coefficients. Learning how the predictors $X$'s regress on the variation of $Y(t)$, as a result, can be naturally replaced with learning how the predictors $X$'s regress on the set of unknown basis coefficients $c$'s. Therefore we propose to set the basis coefficients $c$'s as a function of $X$'s. Let $F(\cdot)$ be a mapping function from $\mathbb{R}^{P}$ to $\mathbb{R}^{K_{b}}$ so that $\boldsymbol{X}$ can be mapped to the basis coefficients as 
\begin{align}
    \boldsymbol{C}_{i} = F(\boldsymbol{X}_{i}),
    \label{F(x)mapping_basiscoef}
\end{align}
which in turn can be used to map $\boldsymbol{X}$ to the functional response $Y(t)$ with $Y_i(t) = \boldsymbol{\theta}'F(\boldsymbol{X}_{i})$. Consequently, we can learn the nonlinear relation between $\boldsymbol{X}_{i}$ and $Y_{i}(t)$ by setting $F(\cdot)$ to be a nonlinear function. Here we propose a dense feed-forward NN as the mapping function $F(\cdot)$, where the basis coefficients $\{c_{i1}, c_{i2},...,c_{iK_{b}}\}$ are the outputs of the NN, then the model can be expressed as
\begin{align}
    \boldsymbol{C}_{i} = \text{NN}_{\eta}(\boldsymbol{X}_{i}) = g_{L+1}\left(\cdots g_{1}\left(\sum_{p=1}^{P}w_{1p}X_{ip}+b_{1}\right)\right),
    \label{NNBB_output_est}
\end{align}
where $g_{1},...,g_{L+1}$ are the activation functions at each layer with $L$ being the number of hidden layers, and $\eta$ denotes the NN parameter set consisting of weights $\{w_{\ell p}\}_{\ell=1}^{L+1}$ and bias $\{b_{\ell}\}_{\ell=1}^{L+1}$ of all layers. The NN is optimized by minimizing the standard objective function calculating the mean squared error (MSE) as $L_{\boldsymbol{C}}(\eta) = 1/n_{\text{train}}
\sum_{i=1}^{n_{\text{train}}}\sum_{k=1}^{K_{b}}(\hat{c}_{ik}-{c}_{ik})^2$, where $n_{\text{train}}$ is the number of samples in the training set, and in application the \textit{true} basis coefficients ${c}_{ik}$'s are replaced with ${c^{\circ}}_{ik}$'s. 

\begin{algorithm}[ht]
\SetAlgoLined
\SetArgSty{}
\setcounter{AlgoLine}{0}
\DontPrintSemicolon
\newcommand{\hrulealg}[0]{\vspace{2mm} \hrule \vspace{1mm}}
\SetKwBlock{Loop}{For}{End}
\linespread{1.1}\selectfont

\KwIn{$\boldsymbol{X}_{i} = \{X_{i1}, X_{i2}, ..., X_{iP}\}$, $\{Z_{i}(t_{ij})\}_{j=1}^{m}$ in the training set $n_{\text{train}}$}
\KwOut {NN with the optimized parameter set $\hat{\eta}$} 
\hrulealg
\nl set up hyper-parameters, including: \\
- for smoothing $Z_{i}(t_{ij})$: basis function vector $\boldsymbol{\theta} = [\theta_1(t), ..., \theta_{K_{b}}(t)]$, number of basis functions $K_{b}$\\
- for training NN: number of hidden layers ($L$), number of neurons per hidden layer, activation functions ($g_1, ..., g_{L+1}$), number of epochs ($E$), batch size, NN optimizer (with a learning rate $\varrho$), etc.

\nl \lForAll{$i \in n_{\text{train}}$}{obtain $\{c^{\circ}_{ik}\}_{k=1}^{K_{b}}$ by projecting $\{Z_i(t_{ij})\}_{j=1}^{m}$ to the selected set of basis functions $\boldsymbol{\theta}$}

\nl randomly initialize NN parameter set and get $\eta = \eta_{\text{initial}}$ 

\nl train a fully-connected NN with input $\boldsymbol{X}_{i}$ and output $\{c^{\circ}_{ik}\}_{k=1}^{K_{b}}$

\For{ $e=1$ \KwTo $E$}{
I. \textit{forward propagation}\newline
\Indp 
(i) pass $\boldsymbol{X}_{i}$ through the NN and get $\{\hat{c}_{ik}\}_{k=1}^{K_{b}}$ for all $i \in n_{\text{train}}$\\
(ii) calculate $L(\eta) = 1/n_{\text{train}}
\sum_{i=1}^{n_{\text{train}}}\sum_{k=1}^{K_{b}}(\hat{c}_{ik}-{c^{\circ}_{ik}})^2$
\Indm

II. \textit{backward propagation}\newline
\Indp
(i) update NN parameter set as $\eta^{\ast} = \eta - \varrho\frac{\partial L(\eta)}{\partial \eta}$
\Indm

III. \lIf{$e < E$}{set $\eta = \eta^{\ast}$, and then repeat I. \& II.}\Indp
\lElse{return $\eta^{\ast}$}
\Indm
}
\nl \KwRet NN with the estimated parameter set $\hat{\eta} = \eta^{\ast}$
\caption{Training NNBB}
\label{NNBB_algorithm}
\end{algorithm}

We named this model NNBB as it is an NN, with B-spline coefficient output (B) and trained by minimizing the MSE of those B-spline coefficients (B). With the exception of the output layer, the NNBB uses a conventional NN architecture and thus, typical approaches for hyper-parameter tuning and training can be applied. We summarize the training process of NNBB in Algorithm \ref{NNBB_algorithm}. The optimized NN is used for prediction where it takes the new scalar inputs in the test set and then outputs the predicted basis coefficients $\hat{\boldsymbol{C}}_{\text{new}}$. The predicted functional response $\hat{\boldsymbol{Y}}(t)$ is further constructed as $\hat{\boldsymbol{Y}}(t) = \boldsymbol{\theta}'\hat{\boldsymbol{C}}_{\text{new}}$. 

\subsection{Neural network with functional response: mapping to FPC scores (NNSS)}
\label{subsec:FPCA}

Apart from basis expansion, the other popular approach for dimension reduction is FPCA. To be specific, let $\mu(t)$ and $K(t, t') = \text{cov}(Y(t), Y(t'))$ be the mean and covariance functions of the underlying function $Y(t)$, and accordingly, the spectral decomposition of the covariance function is $K(t, t') =\sum_{k=1}^{\infty}\gamma_{k}\phi_{k}(t)\phi_{k}(t')$, where $\{\gamma_k, k \geq 1\}$ are the eigenvalues in decreasing order with $\sum_{k}\gamma_k < \infty $ and $\phi_k$'s are corresponding eigenfunctions, also named functional principle components (FPCs), with restriction of $\int\phi_{k}^{2}(t)dt = 1$ for all $k$. Following the Karhunen-Lo\`eve expansion, the $i$-th observed functional response $Y_{i}(t)$ can be represented as $Y_i(t) = \mu(t)+\sum_{k=1}^{\infty}\xi_{ik}\phi_k(t)$. Denote $\Tilde{Y}_i(t) =  Y_i(t) - \mu(t)$ as the centered functional response for the $i$-th subject, we can express the expansion as
\begin{align}
    \Tilde{Y}_i(t) = \sum_{k=1}^{\infty}\xi_{ik}\phi_k(t),
\end{align}
where $\xi_{ik} = \int\{Y_i(t)-\mu(t)\}\phi_k(t)dt$ is the $k$-th FPC score for $Y_i(t)$ with zero mean and $\text{cov}(\xi_{ik}, \xi_{il}) = \gamma_{k}\cdot\mathds{1}(k=l)$. Representing functions with eigenfucntions eigen-decomposed from $\text{cov}(Y(t), Y(t'))$ and the corresponding FPC scores can be considered as a special case of basis expansion, where eigenfunctions play the role of basis functions which, however, are unknown and need to be numerically
calculated using the observed functional data. Given a desired proportion of variance explained ($\tau$) by the FPCs, $\Tilde{Y}_i(t)$ can be approximated arbitrarily well by a finite number of the leading FPCs, where
\begin{align}
    \Tilde{Y}_i(t) \approx \sum_{k=1}^{K_{\tau}}\xi_{ik}\phi_k(t),
    \label{KL}
\end{align}
and $K_{\tau}$ is a valued parameter truncating the FPCs and reducing the dimension of $Y(t)$. $K_{\tau}$ is determined by $\tau$ as the smallest integer satisfying $\sum_{k=1}^{K_{\tau}}\gamma_k \geq \tau$. Because of the fast decay rate of $\gamma_k$, $K_{\tau}$ is usually an small integer. In pratice, FPCA can be easily performed in \texttt{R} with the help of some FDA-related packages, e.g., \textbf{fda} \citep{fda_package} and \textbf{fdapace} \citep{fdapace}. The \textbf{fda} package requires the users to firstly smooth the discrete observations $Z_i(t_{ij})$'s and then apply FPCA on the smoothed functional data for $\{\xi^{\circ}_{ik}\}_{k=1}^{K_{\tau}}$ and $\{\hat{\phi}_k(t)\}_{k=1}^{K_{\tau}}$, an estimator of $\{\xi_{ik}\}_{k=1}^{K_{\tau}}$ and $\{\phi_k(t)\}_{k=1}^{K_{\tau}}$, respectively, while the \textbf{fdapace} package is able to estimate $\{\xi_{ik}\}_{k=1}^{K_{\tau}}$ and $\{\phi_k(t)\}_{k=1}^{K_{\tau}}$ with observed data $Z_i(t_{ij})$'s directly.

It is clearly shown in Eq.(\ref{KL}) that $\Tilde{Y}_i(t)$ can be effectively approximated by a linear combination of the top eigenfunctions $\{\phi_k(t)\}_{k=1}^{K_{\tau}}$, with the major modes of variations among $\Tilde{Y}_i(t)$ captured. Following the idea in Wang \& al. \citep{FoFNN}, under the regression setting, in order to learn the impact of different values of $\boldsymbol{X}_i$ on the variation of $\Tilde{Y}_i(t)$, we propose to set the coefficients $\boldsymbol{\xi}_{i} = \{\xi_{i1}, \xi_{i2},..., \xi_{iK_{\tau}}\}$ to be a function of $\boldsymbol{X}_i$, by formulation, $\boldsymbol{\xi}_{i} = F(\boldsymbol{X}_{i})$ and consequently we have
\begin{align}
     \Tilde{Y}_i(t) = \boldsymbol{\phi}'F(\boldsymbol{X}_i),
     \label{F(x)mapping_centeredFPCA}
\end{align}
where $\boldsymbol{\phi}$ is a $K_{\tau}$-dimensional vector cataloging the $K_{\tau}$ leading FPCs. To model the nonlinear relation, similarly, we proposed to set the mapping function $F(\cdot)$ as a dense feed-forward NN, and then we can write
\begin{align}
      \boldsymbol{\xi}_i = \text{NN}_{\eta}(\boldsymbol{X}_{i}) = g_{L+1}\left(\cdots g_{1}\left(\sum_{p=1}^{P}w_{1p}X_{ip}+b_{1}\right)\right),
    \label{FNN_Y(t)_FPCA}
\end{align}
where $g_{1},...,g_{L+1}$ are the activation functions at each layer with $L$ being the number of hidden layers, and $\eta$ denotes the NN parameter set consisting of weights $\{w_{\ell p}\}_{\ell=1}^{L+1}$ and bias $\{b_{\ell}\}_{\ell=1}^{L+1}$ of all layers. Likewise, $\text{NN}_{\eta}(\boldsymbol{X}_{i})$ is trained by minimizing the MSE loss function $L_{\boldsymbol{\xi}}(\eta)  = 1/n_{\text{train}}
\sum_{i=1}^{n_{\text{train}}}\sum_{k=1}^{K_{\tau}} (\hat{\xi}_{ik}-{\xi}_{ik})^2$, where $n_{\text{train}}$ stands for the number of training set subjects, and the \textit{true} FPC scores $\xi_{ik}$'s are replaced by  $\xi^{\circ}_{ik}$'s. 

\begin{algorithm}[ht] 
\SetAlgoLined
\SetArgSty{}
\setcounter{AlgoLine}{0}
\DontPrintSemicolon
\newcommand{\hrulealg}[0]{\vspace{2mm} \hrule \vspace{1mm}}
\SetKwBlock{Loop}{For}{End}
\linespread{1.1}\selectfont

\KwIn{$\boldsymbol{X}_{i} = \{X_{i1}, X_{i2}, ..., X_{iP}\}$, $\{Z_{i}(t_{ij})\}_{j=1}^{m}$ in the training set $n_{\text{train}}$}
\KwOut {NN with the optimized parameter set $\hat{\eta}$} 
\hrulealg

\nl set up hyper-parameters, including: \\
 - for performing FPCA: the desired proportion of variance explained $\tau$ for determining $K_{\tau}$ \\ 
 - for training NN: number of hidden layers ($L$), number of neurons per hidden layer, activation functions ($g_1, ..., g_{L+1}$), number of epochs ($E$), batch size, NN optimizer (with a learning rate $\tau$), etc.
 
\ForAll{$i \in n_{\text{train}}$}{
 \nl estimate the mean function $\mu(t)$ and the covariance function $K(t, t')$ using $\{Z_i(t_{ij})\}_{j=1}^{m}$
 
 \nl perform eigen-decomposition on $\hat{K}(t, t')$ and get $\{\hat{\phi}_k(t)\}_{k=1}^{K_{\tau}}$ and the corresponding FPC scores $\{\xi^{\circ}_{ik}\}_{k=1}^{K_{\tau}}$
 }
\nl randomly initialize NN parameter set and get $\eta = \eta_{\text{initial}}$

\nl train a fully-connected NN with input $\boldsymbol{X}_{i}$ and output $\{\xi^{\circ}_{ik}\}_{k=1}^{K_{\tau}}$ 

\For{ $e=1$ \KwTo $E$}{

I. \textit{forward propagation}\\
\Indp 
(i) pass $\boldsymbol{X}_{i}$ through the NN and get $\{\hat{\xi}_{ik}\}_{k=1}^{K_{\tau}}$ for all $i \in n_{\text{train}}$\\
(ii) calculate $L(\eta) = 1/n_{\text{train}}
\sum_{i=1}^{n_{\text{train}}}\sum_{k=1}^{K_{\tau}}(\hat{\xi}_{ik}-{\xi}^{\circ}_{ik})^2$
\Indm

II. \textit{backward propagation} \\
\Indp
(i) update NN parameter set as $\eta^{\ast} = \eta - \varrho \frac{\partial L(\eta)}{\partial \eta}$
\Indm

III. \lIf{$e < E$}{set $\eta = \eta^{\ast}$, and then repeat I. \& II.}
\Indp
\lElse{return $\eta^{\ast}$}
\Indm
}
\nl \KwRet NN with the estimated parameter set $\hat{\eta} = \eta^{\ast}$ 

\caption{Training NNSS}
\label{NNSS_algorithm}
\end{algorithm}

Following the same naming convention, we refer to this model has NNSS because it is an NN model build with output being the FPC scores (S) trained on the scores (S) themselves.The complete training process of NNSS is summarized in Algorithm \ref{NNSS_algorithm}. Again, the hyper-parameter tuning and model training of NNSS are the same as of a conventional NN. Meanwhile, the returned NN with optimized parameters set $\hat{\eta}$ is used for prediction where it takes the new scalar inputs in the test set and then outputs the predicted FPC scores $\hat{\boldsymbol{\xi}}_{\text{new}}$. The predicted functional response $\hat{\boldsymbol{Y}}(t)$ is further recovered as $\hat{\boldsymbol{Y}}(t) = \hat{\mu}(t) + \hat{\boldsymbol{\phi}}'\hat{\boldsymbol{\xi}}_{\text{new}}= \hat{\mu}(t) + \hat{\boldsymbol{\phi}}' \text{NN}(\boldsymbol{X}_{\text{new}}|\hat{\eta})$, where $\hat{\boldsymbol{\phi}}$ is the vector consisting of the estimated FPCs $\hat{\phi}_1(t), ..., \hat{\phi}_{K_{\tau}}(t)$.

\subsection{Modification to the objective function (NNBR and NNSR)}

In the two previous sections, we define two NNs outputting basis coefficients or FPC scores, and those outputs are further used to construct the predicted response variable. Let us discuss a way to build an objective function that directly uses the response variable in order to estimate the parameters. For simplicity, we only discuss the B-spline model of Section \ref{subsec:be} but know that a similar concept can also be applied to the FPCA model described in Section \ref{subsec:FPCA}. In brief, in Section \ref{subsec:be}, we first fit a B-spline model on the observed functional response to estimate a set of basis coefficients for the response in the training set and then we train an NN to predict these basis coefficients using the predictors. 

In this section, we propose to modify the objective function in order to bypass the initial estimation of basis coefficients. The key idea of the new objective function is to directly minimize the prediction error of the response variable. In other words, instead of minimizing the MSE between $\hat{c}_{ik}$ and $c_{ik}$, here we first transform the NN output, the B-splines basis coefficients, into the predicted response $\hat{Y_i}(t)$ and then minimize the MSE between $\hat{Y_i}(t)$ and $Y_i(t)$. What supports us to implement such objective function is the fact that we rely on differentiable operations to build this new objective function. Doing so guarantees that we can rely on the back-propagation algorithm to train the NN using readily available packages. 

Suppose that we want to fit the functional response with a B-spline made of $K_{b}$ basis functions $\boldsymbol{\theta}$ and $K_{b}$ associated basis coefficients $\boldsymbol{C}$ . Then the predict response $\hat{Y}_i(t)$ at time $t$ is the vector product between $\boldsymbol{\theta}$ and $\boldsymbol{C}_{i}$ as illustrated in Eq. (\ref{BasisExpansion_EQ}). This means that the relation between the predicted response $\hat{Y}_i(t)$ and the predicted basis coefficients $\boldsymbol{\hat{C}}_{i}$ is linear, thus we can easily compute the derivative of $\hat{Y}_i(t)$ with respect to the coefficients. This further indicates that if we observe the response at time $t$ then we are able to compute the gradient of $(Y_i(t)-\hat{Y}_i(t))^2$ with respect to the basis coefficients and therefore we can also compute the gradient with respect to the parameters $\eta$ of the NN that outputs those basis coefficients.

More generally, assuming the response is observed at $m$ time points $t_1,t_2,...,t_m$ for every subject, we can generate a matrix of basis functions evaluated at those time points, denoted as $\Theta$, which is a $K_{b} \times m$ matrix where each row represents a basis function with entries taking the values of that basis function at each of the $m$ time points. Consequently, we can obtain the predicted response $\hat{Y}_{(n_{\text{train}} \times m)}$ for every $m$ observed time points and for all $n_{\text{train}}$ subjects by doing a simple matrix multiplication
\begin{align}
     \hat{Y}_{(n_{\text{train}} \times m)} = \hat{\boldsymbol{C}}_{(n_{\text{train}}\times K_{b})}\Theta_{(K_{b}\times m)},
     \label{matrix_multiplication}
\end{align}
where $\hat{\boldsymbol{C}}$ is a matrix where a row contains the $K_{b}$ NN-estimated coefficients for a single observed subject. In our proposed model, $\hat{\boldsymbol{C}}$ is the output produced by the NN function as explained in Section \ref{subsec:be}. However, we now modify the objective function to train the NN to minimize the MSE between the observed response and the predicted one
\begin{align}
    L_{\boldsymbol{Y}}(\eta) = \frac{1}{n_{\text{train}}}
    \sum_{i=1}^{n_{\text{train}}}\sum_{j=1}^{m}(Y_{i}(t_{ij})-\hat{Y}_{i}(t_{ij}))^2. 
\end{align}
Note that in implementation, the discrete observation $Z_{i}(t_{ij}) = Y_{i}(t_{ij}) + \epsilon_{i}(t_{ij})$ will replace $Y_{i}(t_{ij})$, the underlying \textit{true} functional curve. In this way, we entirely bypass the need to estimate the basis coefficients first and fit the parameters $\eta$ with respect to the functional response directly. We call this variant NNBR because its architecture is an NN with basis coefficients output (B) fitted by minimizing the MSE of the response variable (R). The entire training process of NNBR is detailed in Algorithm \ref{NNBR_algorithm}. Similarly, we can get a prediction for a new input $\boldsymbol{X}_{\text{new}}$ as $\hat{Y}(t) = \boldsymbol{\theta}'\hat{\boldsymbol{C}}_{\text{new}}=\boldsymbol{\theta}' \text{NN}(\boldsymbol{X}_{\text{new}}|\hat{\eta})$.

\begin{algorithm}[ht]
\SetAlgoLined
\setcounter{AlgoLine}{0}
\SetArgSty{}
\DontPrintSemicolon
\newcommand{\hrulealg}[0]{\vspace{2mm} \hrule \vspace{1mm}}
\SetKwBlock{Loop}{For}{End}
\linespread{1.15}\selectfont

\KwIn{$\boldsymbol{X}_{i} = \{X_{i1}, X_{i2}, ..., X_{iP}\}$, $\{Z_{i}(t_{ij})\}_{j=1}^{m}$ in the training set $n_{\text{train}}$}
\KwOut {NN with the optimized parameter set $\hat{\eta}$} 
\hrulealg

\nl set up hyper-parameters, including: \\
 - for smoothing $Z_{i}(t_{ij})$: basis function vector $\boldsymbol{\theta} = [\theta_1(t), ..., \theta_{K_{b}}(t)]$, number of basis functions $K_{b}$\\
 - for training NN: number of hidden layers ($L$), number of neurons per hidden layer, activation functions ($g_1, ..., g_{L+1}$), number of epochs ($E$), batch size, NN optimizer (with a learning rate $\varrho$), etc.
 
\nl evaluate $[\theta_1(t), ..., \theta_{K_b}(t)]$ at all observed timestamps $\{t_j\}_{j=1}^{m}$ and form the matrix $\Theta_{(K_{b} \times m)}$

\nl randomly initialize NN parameter set and get $\eta = \eta_{\text{initial}}$

\nl train a fully-connected NN with input $\boldsymbol{X}_{i}$ and output $\{Z_i(t_{ij})\}_{j=1}^{m}$

\For{ $e=1$ \KwTo $E$}{

I. \textit{forward propagation}\\
\Indp 
(i) pass $\boldsymbol{X}_{i}$ through the NN and get $\{\hat{c}_{ik}\}_{k=1}^{K_{b}}$ for all $i \in n_{\text{train}}$\\
(ii) multiply $\{\hat{c}_{ik}\}_{k=1}^{K_{b}}$ with $\Theta_{(K_{b} \times m)}$ to get $\{\hat{Y}(t_{ij})\}_{j=1}^{m}$  \\
(iii) calculate $L(\eta) = 1/n_{\text{train}}
    \sum_{i=1}^{n_{\text{train}}}\sum_{j=1}^{m}(Z_{i}({ij})-\hat{Y}_{i}(t_{ij}))^2$
\Indm

II. \textit{backward propagation} \\
\Indp
(i) update NN parameter set as $\eta^{\ast} = \eta - \varrho\frac{\partial L(\eta)}{\partial \eta}$
\Indm

III. \lIf{$e < E$}{set $\eta = \eta^{\ast}$, and then repeat I. \& II.}
\Indp
\lElse{return $\eta^{\ast}$}
\Indm
}
\nl \KwRet NN with the estimated parameter set $\hat{\eta} = \eta^{\ast}$

\caption{Training NNBR}
\label{NNBR_algorithm}
\end{algorithm}

When comparing NNBR to NNBB, we note that they share a similar architecture but use different objective functions. A significant difference between both models is that in order to train the NNBB we need to first estimate the basis coefficients of the functional response. Conversely, training the NNBR requires a modification of the objective function which slightly increases the computational cost but meanwhile bypasses the need of a beforehand estimation of basis coefficients.

A similar process could be conducted with an NN that outputs FPC scores as described in Section \ref{subsec:FPCA}, where this time it predicts response $\hat{Y}$ using those scores. We named that model NNSR and describe the training process in Algorithm \ref{NNSR_algorithm}.

It is worth mentioning that the network architectures of NNBR and NNSR remain the same as a classic NN even though we have modified their objective functions. Such modifications only influence the network optimization with no changes to the sensitivity with respect to the hyper-parameters and the architecture (number of layers and depth). Additionally, the proposed modification that targets at the minimization of the difference between $Y_i(t)$ and $\hat{Y}_i(t)$ is surprisingly simple, convenient and computationally efficient. Compared to the operation that minimizes the difference between $c_{ik}$ and $\hat{c}_{ik}$ (NNBB) or between $\xi_{ik}$ and $\hat{\xi}_{ik}$ (NNSS), the modified objective function only requires an additional operation, namely a simple matrix multiplication as Eq. (\ref{matrix_multiplication}) in forward propagation and an additional step to compute the gradient of $(Y_i(t)-\hat{Y}_i(t))^2$ with respect to the basis coefficients outputted by NN in backward propagation. This only increases negligibly the computational cost of NNBR and NNSR when compared to NNBB or NNSS.

\begin{algorithm}[ht] 
\SetAlgoLined
\setcounter{AlgoLine}{0}
\SetArgSty{}
\DontPrintSemicolon
\newcommand{\hrulealg}[0]{\vspace{2mm} \hrule \vspace{1mm}}
\SetKwBlock{Loop}{For}{End}
\linespread{1.15}\selectfont

\KwIn{$\boldsymbol{X}_{i} = \{X_{i1}, X_{i2}, ..., X_{iP}\}$, $\{Z_{i}(t_{ij})\}_{j=1}^{m}$ in the training set $n_{\text{train}}$}
\KwOut {NN with the optimized parameter set $\hat{\eta}$} 
\hrulealg

\nl set up hyper-parameters, including: \\
 - for performing FPCA: the desired proportion of variance explained $\tau$ for determining $K_{\tau}$ \\ 
 - for training NN: number of hidden layers ($L$), number of neurons per hidden layer, activation functions ($g_1, ..., g_{L+1}$), number of epochs ($E$), batch size, NN optimizer (with a learning rate $\tau$), etc.
 
\ForAll{$i \in n_{\text{train}}$}{

 \nl estimate the mean function $\mu(t)$ and the covariance function $K(t, t')$ using $\{Z_i(t_{ij})\}_{j=1}^{m}$ 
 
 \nl perform eigen-decomposition on $\hat{K}(t, t')$ and get $\{\hat{\phi}_k(t)\}_{k=1}^{K_{\tau}}$
 
 \nl form $\hat{\Phi}_{(K_{\tau} \times m)}$, a matrix of eigenfunctions $[\hat{\phi}_1(t), ..., \hat{\phi}_{K_\tau}(t)]$ evaluated at all observed time points $\{t_j\}_{j=1}^{m}$
 }
\nl randomly initialize NN parameter set and get $\eta = \eta_{\text{initial}}$

\nl train a fully-connected NN with input $\boldsymbol{X}_{i}$ and output $\{Z_i(t_{ij})\}_{j=1}^{m}$ 

\For{ $e=1$ \KwTo $E$}{

I. \textit{forward propagation}\\
\Indp 
(i) pass $\boldsymbol{X}_{i}$ through the NN and get $\{\hat{\xi}_{ik}\}_{k=1}^{K_{\tau}}$ for all $i \in n_{\text{train}}$\\
(ii) multiply $\{\hat{\xi}_{ik}\}_{k=1}^{K_{\tau}}$ with $\hat{\Phi}_{(K_{\tau} \times m)}$ to get $\{\hat{Y}(t_{ij})\}_{j=1}^{m}$ \\
(iii) calculate $L(\eta) = 1/n_{\text{train}}
    \sum_{i=1}^{n_{\text{train}}}\sum_{j=1}^{m}(Z_{i}({ij})-\hat{Y}_{i}(t_{ij}))^2$
\Indm

II. \textit{backward propagation} \\
\Indp
(i) update NN parameter set as $\eta^{\ast} = \eta - \varrho \frac{\partial L(\eta)}{\partial \eta}$
\Indm

III. \lIf{$e < E$}{set $\eta = \eta^{\ast}$, and then repeat I. \& II.}
\Indp
\lElse{return $\eta^{\ast}$}
\Indm
}
\nl \KwRet NN with the estimated parameter set $\hat{\eta} = \eta^{\ast}$

\caption{Training NNSR}
\label{NNSR_algorithm}
\end{algorithm}

\subsection{Irregularly-spaced functional data}

Earlier in this section we made the common assumption that the functional response $Y(t)$ is observed at the same $m$ equally-spaced time points. While this was a useful assumption to make in order to explain how to train the model, it is actually not a necessary condition to fit any of the four models described above and we can train these models even with irregularly spaced functional response. 

For the first two models we introduced, NNBB and NNSS, in order to train the NN, we simply need an estimate for the basis coefficients or an estimate for the FPC scores. We can rely on some existing FDA literature to get those estimates \citep{fda,yao2005functional} for irregularly-spaced functional data.

For both models that utilize the modified objective function, NNBR and NNSR, it is a bit more complicated but not so much. Once again, let us focus on NNBR to simplify the explanations. In the setting with irregularly-spaced functional observations, the assumptions $m_1 = m_2 = \cdots = m_N = m$ and $t_{1j} = t_{2j} = \cdots = t_{Nj} = t_j$ no longer hold. Suppose that $m_{\text{irr}}$ is the total number of time points with at least one observation given all training subjects and $\{t_{ij}\}_{j=1}^{m_{\text{irr}}}$ represents the union set of $\{t_{ij}\}_{j=1}^{m_i}$ for all $i$ in the training set. The goal is to train the model using only the observations at $\{t_{ij}\}_{j=1}^{m_i}$ when the $i$-th subject is observed at this time. To achieve this, we need to generate a matrix $\Theta_{\text{irr}}$ of size $K_b \times m_{\text{irr}}$ with the $k$-th row containing entries of the $k$-th chosen basis function evaluated at $\{t_{ij}\}_{j=1}^{m_{\text{irr}}}$. This matrix $\Theta_{\text{irr}}$ will be employed for the calculation of $\hat{Y}_i(t_{ij})$ at all $m_{\text{irr}}$ time points with observation(s) by taking the matrix multiplication $ \text{NN}_{\eta}(\boldsymbol{X})\Theta_{\text{irr}}$. However, because for each subject $i$, $Y_i(t_{ij})$'s are originally observed at only $\{t_{ij}\}_{j=1}^{m_{i}}$ instead of all $m_{\text{irr}}$ time points, when computing the objective function, we need to eliminate the contribution of those unobserved subject-time point pairs by multiplying them by 0, and accordingly
\begin{align}
    L_{\boldsymbol{Y_{\text{irr}}}}(\eta) & = \frac{1}{n_{\text{train}}}\sum_{i=1}^{n_{\text{train}}}\sum_{j=1}^{m_{i}}\left(Y_{i}(t_{ij})-\hat{Y}_{i}(t_{ij})\right)^{2} \notag \\
    & = \frac{1}{n_{\text{train}}}\sum_{i=1}^{n_{\text{train}}}\sum_{j=1}^{m_{\text{irr}}}\left(Y_{i}(t_{ij})-\hat{Y}_{i}(t_{ij})\right)^{2}\cdot\mathds{1}\left(Y_{i}(t_{ij})\text{ is observed}\right),
\end{align}
which assures that the unobserved subject-time point pairs do not contribute to the gradient of the objective function.

Because of the common strategy used in FDA, which represents the infinite-dimensional curve using a finite basis set, we are able to construct the curve over the entire interval $\mathcal{T}$. This allows us to compute the predictive accuracy even if the test set contains observations at time points previously unseen, as long as they are within the interval of time points observed in the training set. 

\subsection{Roughness penalty}
When predicting the functional response using NNs, we need to pay attention not only to the predictive accuracy but also the smoothness of the predicted trajectories, as it is standard in FDA. In practice, we observe that the performance of the NN mapping with basis expansion is influenced by the number of basis functions selected. Usually, the more basis functions we use to re-express the functional response, the higher accuracy can be gained when predicting the response at a set of given time points. 

In FDA, it is common to set the number of basis functions to be less than the number of the observed time/location points with enough valid observations. However, in the cases where the observed time points are limited, we can consider increasing the number of basis functions to benefit the predictive accuracy. This benefit brought by more basis functions comes at the sacrifice of smoothness of the fitted functional trajectories, as they introduce in more variations. This is to be expected; because the prediction response for a specific time point $\hat{Y_i}(t)$ is the inner product of the basis coefficients and basis functions that are non-zero at $t$, then the more non-zero basis functions we have the more flexible that point estimate is. To control the smoothness of fitted curves without limiting the number of basis functions,  we follow the common idea in FDA and propose to add some classic roughness penalty to the objective function of the NN. This simple addition ensures the smoothness of the predicted functional curves without putting any burden to the differentialable operation (since the roughness penalty terms we consider are also linearly related to the NN outputs).

There are multiple types of roughness penalties that can be applied to smooth the functional curves. In our implementation, we focus on two conventional and popular approaches: penalizing with the second derivative of the function, and penalizing directly the basis coefficients, to relieve the roughness concern with the NN-fitted functional response. Notice that we do not need to define roughness penalties for NNBB and NNSS as the smoothing is done when fitting the curve a priori.

\subsubsection{Penalizing the second-order derivative of \texorpdfstring{$Y(t)$}{Y(t)}}

When the smoothness of fit becomes a concern and the roughness penalty turns out to be a necessity, the most common and popular method would be penalizing the $o$-th order derivative of the function $Y(t)$. The squared second derivative of a function $Y(t)$ at $t$ reveals its curvature at $t$, therefore it is natural to measure the roughness of the $\hat{Y}_{i}(t)$ by taking the integrated square of its second derivative \citep{fda}. Adding this penalty term to the objective function of the NN leads to
\begin{align}
     L_{\boldsymbol{pen}}(\eta) = \frac{1}{n_{\text{train}}}
    \sum_{i=1}^{n_{\text{train}}}\left(\sum_{j=1}^{m}\left(Y_{i}(t_{ij})-\hat{Y}_{i}(t_{ij})\right)^2+\lambda\int_{\mathcal{T}}\left(\frac{d^{2}\hat{Y}_{i}(t)}{dt^{2}}\right)^{2}dt\right).
    \label{penalty_2nd.deriv}
\end{align}
The parameter $\lambda$ acts as a smoothness controller for balancing the trade-off between fitting to the data and the variability of the predicted function \citep{fda}. When $\lambda$ becomes larger, the more emphasis will be put on the smoothness of the fitted curves. On the contrary, for a smaller $\lambda$, the fitted curve tends to be more wiggly as there is less penalty placed on its roughness. The selection of the optimal $\lambda$ can be achieved by using cross-validation, where either the subjects or the time points are randomly divided into sub-samples (in practice, we partition the time points).

Unfortunately, we cannot back-propagate the gradient through the integral of Eq. (\ref{penalty_2nd.deriv}), and thus we will approximate the integral with a summation over the domain $\mathcal{T}$. Because we are able to generate $\hat{Y}_i(t)$, we can approximate this integral with as many points as we deemed necessary
\begin{align}
    L_{\boldsymbol{pen}}(\eta) = \frac{1}{n_{\text{train}}}
    \sum_{i=1}^{n_{\text{train}}}\left(\sum_{j=1}^{m}(Y_{i}(t_{ij})-\hat{Y}_{i}(t_{ij}))^2+\frac{\lambda T}{Q-1}\sum_{q=2}^{Q}\left(\frac{d^{2}\hat{Y}_{i}(t_{iq})}{dt_{iq}^{2}}\right)^{2}\right),
    \label{penalty_2nd.deriv_simplify}
\end{align}
where $t_{iq} = t_{q}$ for all $i$ and $\{t_{q}\}_{q=1}^{Q}$ are equally spaced time points covering the entire domain $\mathcal{T}$ with length $T$ . Notice that we do not need to actually compute those second order derivatives, because
\begin{align}
\hat{Y}_i(t) &= \sum_{k=1}^{K_{b}}\hat{c}_{ik}\theta_{k}(t) \notag \\ 
\Rightarrow \frac{d^{2}\hat{Y}_{i}(t)}{dt^{2}} &= \sum_{k=1}^{K_{b}}\hat{c}_{ik} \frac{d^{2}\theta_{k}(t)}{dt^{2}},
\end{align}
then what we really need are the second order derivatives of the basis functions, which are much easier to compute and readily available in the \textbf{fda} package. Therefore, we can back-propagate the gradient of the objective function of Eq. (\ref{penalty_2nd.deriv_simplify}) with respect to $\eta$ without any problems.

\subsubsection{Penalizing the basis coefficients \texorpdfstring{$C$}{Ci}}
Penalizing directly on the basis coefficients was firstly introduced by Eilers \& Marx \citep{Eilers&Marx} when adjacent B-splines are used to re-express the functional variable in a regression problem. Compared to penalizing the derivative, this idea reduces the dimensionality of the smoothing problem to $K_b$, the number of basis functions, instead of $N$, the number of observations \citep{Eilers&Marx}. 

The basic structure of this penalty is the difference of a set of consecutive basis coefficients $\Delta^{2}c_{k} = c_{k}-2c_{k-1}+c_{k-2}$. This difference has a strong connection with the second derivative of the fitted function, which is revealed by the simple formula for derivatives of B-splines given by De Boor \citep{deboor}. The summation of the squared differences for all three consecutive coefficients $\{c_{k}, c_{k-1}, c_{k-2}\}, k = 3, ..., K_b$ would be the main component of the penalty term, with the strength of the penalty further controlled by a tuning parameter $\lambda$. Like previously, the penalty term is added to the objective function of NN as:
\begin{align}
    L_{\boldsymbol{pen}}(\eta) = \frac{1}{n_{\text{train}}}
    \sum_{i=1}^{n_{\text{train}}}\left(\sum_{j=1}^{m}\left(Y_{i}(t_{ij})-\hat{Y}_{i}(t_{ij})\right)^2+\lambda\sum_{k=3}^{K_b}(\Delta^{2} c_{ik})^{2}\right).
    \label{penalty_basis.coef}
\end{align}
Similarly, the optimal hyper-parameter $\lambda$ is selected using $k$-fold cross-validation.

Empirically, when fitting one of the B-spline models, especially the NNBR model, including a large number of basis functions, more than $m$ for example, leads to a better predictive accuracy, but also results in a set of very rough predicted curves. When $K_b > m$ the predicted curves quickly become \textit{wiggly}, and thus we would increase the number of basis such that $K_b >> m$ and impose a roughness penalty at the same time to keep the curves smooth. In some cases, this leads to the top performer in terms of predictive accuracy. Unfortunately, using $K_b >> m$ and a roughness penalty could make the complex hyper-parameter tuning process even more difficult. Nonetheless, we believe it is important to provide roughness penalties and justifications for those when providing a technique to fit functional data.

We implemented the four models and the two roughness penalties in \texttt{R} \citep{R}. The functional component of our implementation relies on the \textbf{fda} package and the NN component uses the \texttt{R} implementation of \textbf{Keras}. Our implementations of those models along the real data example are available online on the second author's GitHub page.

\section{Computational complexity}\label{sec:computationalcost}

An advantage of using an NN as the link function instead of a linear combination is that it might have a lower computational cost. More specifically, when using a standard training procedure for both models, the NN approach scales better with the number of predictors. Indeed, we usually find an exact solution for the FoS model which involves the inversion of a matrix, resulting in worse scaling with respect to the number of predictors. On the flip side, the algorithm used to train NN scales linearly with the number of predictors. 

Start by looking at the FoS model, we rely on the formulation established in Ramsay \& Silverman \citep{fda}, chapter 14. This book contains various information about the computational details of the solutions of the FoS optimization problem under various parameterizations. However, its simplest form is quite similar to the least square solution of a traditional regression where:
\begin{align}
\mathbf{\hat{B}} = (\mathbf{X}^{T}\mathbf{X})^{-1}\mathbf{X}^TY,
\end{align}
which involves the inversion of a $P\times P$ matrix that requires a number of operations proportional to $P^3$, say $O(P^3)$ in big O notation, using Gaussian elimination. The Strassen algorithm \citep{strassen1969gaussian} manages to get that order down to $O(P^{2.8})$ and the more recent Coppersmith-Winograd algorithm \citep{coppersmith1987matrix} gets the order of matrix multiplication down to $O(P^{2.37})$ for a matrix of dimension $P \times P$. 

The \textbf{fda} package uses the \texttt{solve()} function included in the default \texttt{R} language which in turn employs the linear algebra package (LAPACK) \citep{lapack99} that relies one Gaussian elimination to inverse matrices. Other than the matrix inversion, the matrix multiplication component of the solution involves a $P \times P$ matrix, a $P \times N$ matrix and a $N \times m$ matrix resulting in a number of operations of order $O(PN)$. Including the matrix multiplication part of the solution the whole FoS estimation requires a number of operations of order $O(P^3N)$. The more complete solutions that consider the smoothness of the predicted curves involve the inversion of a matrix of dimension $K_pP \times K_pP$ where $K_p$ is the number of basis of the functional parameters $\boldsymbol{\beta}(t)$. It means scaling polynomial with power 3 with the number of predictors $P$ when using Gaussian elimination. Hence, we consider that the number of operations needed to obtain the exact solution to the FoS problem requires a number of operations of order $O(P^3N)$.  

As far as NNs are concerned, according to Hastie \& al. \citep{hastie2009elements}, the computational cost of training an NN is of order $O(PNn_{w}E)$ when back-propagation is adopted \citep{rumelhart1986} to estimate the gradient and when a gradient-based approach is applied to fit the NN. In the formulation above, $N$ is the number of observations, $P$ is the number of predictors, $n_w$ is the number of weights (hidden neurons) in the NN and $E$ the number of training epochs.

Therefore, if we strictly focus on how the run time of an NN scales with $N$ and $P$, we are looking at a linear scaling in both cases, $O(PN)$. Consequently, when comparing computational costs, the NN model has a pretty significant advantage when it comes to its scaling with respect to the number of predictors. Because of this better scaling with respect to the number of predictors, we can claim that our proposed models are better equipped than the FoS regression to deal with data sets containing lots of predictors.

However, let us be a little more nuanced. It is clear that under some circumstances we can fit extremely large NNs which could lead to a slow fitting process. Additionally, the main difference between the traditional ways to fit these models is that we pursue an exact solution for the linear problem and on the contrary we utilize a gradient-based approach to find a solution in the case of NNs. As a result, for a linear model, such as FoS, we do have worse scaling with respect to $P$ but we have guarantees that the optimal solution is reached. In theory, it would be possible to fit a linear model, such as FoS, with a gradient-based approach to improve how its computational cost scales with respect to $P$, though this is certainly not the standard optimization approach.

\section{Real data application}\label{realapplication}

We evaluate the predictive performance of the proposed methods, along with the conventional FoS model and the FAM model on the ASFR data set introduced in Section \ref{sec:intro}. We consider to compare the proposed models with the FAM model, $Y_i(t) = \sum_{p=1}^{P}f_{p}\left(X_{ip}, t\right) + \epsilon_i(t)$, because our methods can easily and naturally account for the nonlinear effects of all possible interactions simultaneously, while FAM is predominantly focused on learning the nonlinear relation between the functional response and each of the scalar predictors individually (it is computationally difficult and expensive for FAM to learn all possible interactions from a multi-dimensional aspect). For each of the models, we proceed with 20 repetitions of random subsampling validation: randomly dividing the data set into a training set and a test set, with 80\% and 20\% of the total samples assigned to them, respectively. 

Following the same tuning strategy usually applied in classic NNs, the hyper-parameters of all models (except FAM) in comparison are firstly tuned using $5$-fold cross-validation in order to fairly improve their performances during actual training. We start the hyper-parameter tuning with FoS, considering it involves only one hyper-parameter $K_b$, the number of basis functions. To reduce the computational time consumed by the tuning processes of the NN-based models, the optimal basis system selected for the FoS model is then used for the rest of the models. Afterwards, for each NN-based model, we perform a grid search on all network hyper-parameters simultaneously by taking a list of possible values for each of the hyper-parameters and running a $5$-fold cross-validation for all combinations. An NN consisting of 2 hidden layers with 50 and 30 neurons, respectively, is selected. 99\% proportion of variance explained is recommended by cross-validation as the threshold to truncate the FPCs scores in NNSS and NNSR. For NNBR with penalized objective function (NNBR(P)), we apply 10 basis functions and the second-order derivative roughness penalty, together with smoothing parameter $\lambda = 10^{-7}$, which is similarly suggested by the 5-fold cross-validation from a set of possible values $\{10^{-i}\}_{i=1}^{8}$. Considering the number of basis functions is not dramatically large, the fitted curves are already quite smooth without any roughness penalty. Thus, a relatively small $\lambda$ is acceptable as more emphasis ought to be placed on the fitting to data for an overall smallest $L_{\boldsymbol{pen}}(\eta)$. A summary of the optimal hyper-parameters determined for each model is provided in Table S2 in the supplementary document. This hyper-parameter tuning approach is also applied in the simulations of Section \ref{simulation}. The FoS model used in this experiment is coming from the \texttt{R}-package \textbf{fda}, and FAM model is trained with the help of function \texttt{pffr()} in the \textbf{refund} package.

\begin{table}[ht]
\begin{minipage}{\textwidth}
\centering
\caption{Table of MSEPs of 20 random test sets for various models with ASFR data set.}
\label{table:asfr_result}
\begin{tabular}{@{}c c c c@{}}
\toprule
Methods & Mean & Std. Dev. & $p$-value of $t$-test \\ 
\midrule
FoS & 1177.98 & 513.58 & - \\
NNBB & 1031.87 & 294.87 & 0.18\\
NNSS & \textbf{992.50} & 296.68 & \textbf{0.01} \\
NNBR & 1060.50 & 210.63 &  0.20\\
NNBR(P) & 1059.07& \textbf{208.27} & 0.19\\
NNSR & 1047.46 & 277.37 & 0.11 \\
FAM & 1317.33 & 919.62 & 0.28\\
\botrule
\end{tabular}
\end{minipage}
\end{table}
Table \ref{table:asfr_result} shows the predictive performance of each model using the mean squared error of prediction (MSEP) over the 20 replications. The predictive accuracy is measured by the MSEP averaged across the number of samples and the number of observed time points in the test set. A two-sided paired $t$-test is later conducted to compare the MSEPs of the 20 replicates of our models to the 20 MSEPs of the FoS model. It is clearly shown that our proposed models consistently outperform FoS and FAM models in predicting the fertility curves, with both smaller means and standard deviations (SDs) of the prediction errors of 20 random test sets, demonstrating their advantages in capturing both linear and nonlinear effects of all predictors on the functional response. NNSS model, as indicated in Table \ref{table:asfr_result}, has the best predictive performance, which interestingly implies that the FPC scores are more informative than the basis coefficients in representing $Y(t)$ with this data set, and consequently the effects of covariates on functional response are better described by the relations between the predictors and FPC scores instead of basis coefficients. In addition, the prediction error of NNBR optimized by the penalized objective function is lower than that of the non-penalized NNBR, indicating that including a relatively large number of basis functions, along with a roughness penalty term added to the objective function, can help improve the prediction without loss on the smoothness of the predicted curves. 

\begin{figure}[ht]
    \centering
    \includegraphics[width=0.9\columnwidth]{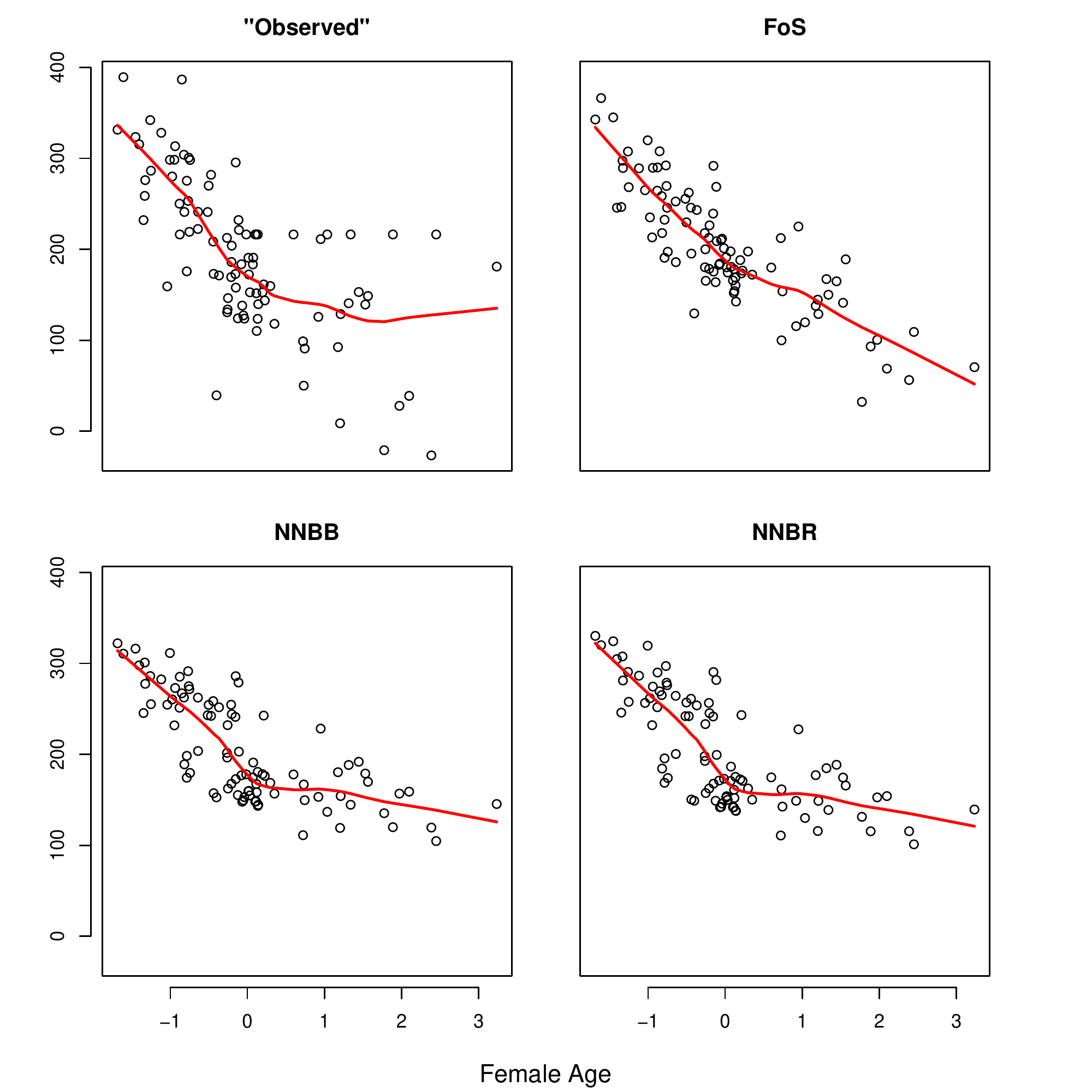}
    \caption{A comparison of the relations between the covariate female age and the $Y(t)$-estimated second basis coefficient ($\hat{c}_{2, Y(t)}$), FoS-predicted second basis coefficient ($\hat{c}_{2, \text{FoS}}$), NNBB-predicted second basis coefficient ($\hat{c}_{2, \text{NNBB}}$) and NNBR-predicted second basis coefficient ($\hat{c}_{2, \text{NNBR}}$), respectively.}
    \label{fig:asfr_basis_vs_femaleage}
\end{figure}
\begin{figure}[ht]
    \centering
    \includegraphics[width=0.9\columnwidth]{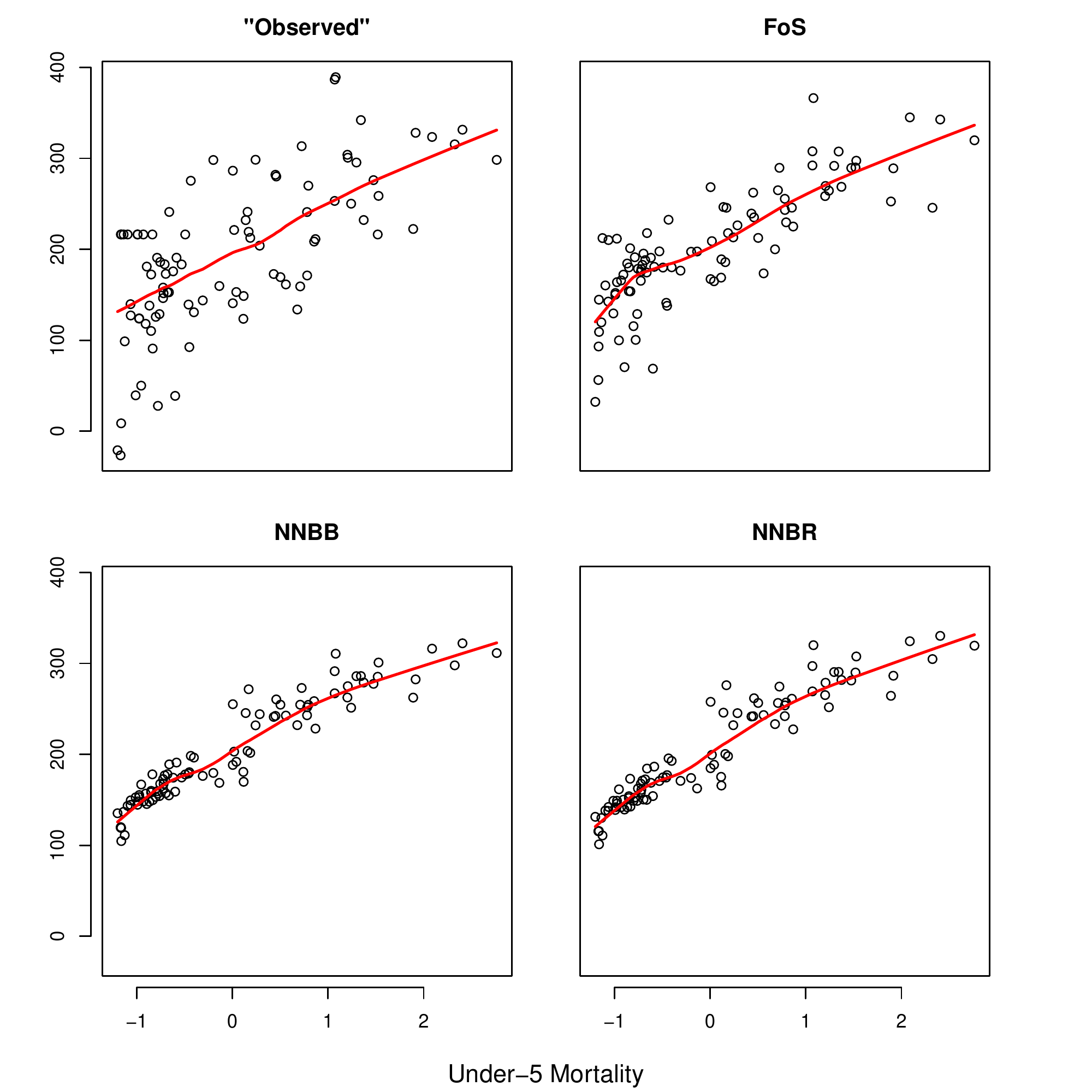}
    \caption{A comparison of the relations between the covariate Under-5 Mortality and the $Y(t)$-estimated second basis coefficient ($\hat{c}_{2, Y(t)}$), FoS-predicted second basis coefficient ($\hat{c}_{2, \text{FoS}}$), NNBB-predicted second basis coefficient ($\hat{c}_{2, \text{NNBB}}$) and NNBR-predicted secondbasis coefficient ($\hat{c}_{2, \text{NNBR}}$), respectively.}
    \label{fig:asfr_basis_vs_u5}
\end{figure}

As previously mentioned in Section \ref{sec:intro}, several covariates, such as female age and GDP per capita, are seemingly not linearly related to many of the basis coefficients representing the fertility trajectory. Therefore we focus on the basis coefficients $\hat{\boldsymbol{C}}_{\text{FoS}}$, $\hat{\boldsymbol{C}}_{\text{NNBB}}$ and $\hat{\boldsymbol{C}}_{\text{MMBR}}$ predicted by FoS, NNBB and NNBR (they are the models using basis coefficients for regression), respectively. The relations between the predictor female age and each of the obtained second basis coefficients, including $Y(t)$-estimated second basis coefficient $\hat{c}_{2, Y(t)}$ and the model-predicted ones, are displayed in Figure \ref{fig:asfr_basis_vs_femaleage}. We can see that the nonlinear pattern for female age and the second basis coefficient is better recovered by NNBB and NNBR, while the FoS-predicted basis coefficient $\hat{c}_{2, \text{FoS}}$ is more likely linearly related to female age. 
Likewise, we select under-5 mortality out of those covariates that have highly likely linear associations with some of the basis coefficients, and Figure \ref{fig:asfr_basis_vs_u5} reveals how under-5 mortality relates to different model-predict second basis coefficient $\hat{c}_{2}$. It is noticed there are plainly linear trends between under-5 mortality and the second basis coefficients predicted by the three models mentioned, while under-5 mortality has been shown linearly related to the second basis coefficient estimated by $Y(t)$. It is not surprising that FoS reconstructs the most similar pattern to the one built on the $Y(t)$-estimated basis coefficient, but the proposed models also master the linear information when recovering the associations, which proves that the NN-based models have the ability to take good care of both the nonlinear and linear relations between the predictors and scalar representations simultaneously. Due to the relatively small number of observations, it is worth mentioning that the relations between covariates and basis coefficients may not be fully revealed and visual misunderstanding could occur.

\section{Simulation studies}\label{simulation}

\subsection{Generating data} \label{generator}

We are interested in data sets that have multiple predictors and a potentially nonlinear relation between the scalar predictors and the functional response.
 We were inspired by the variable selection literature to design our data generators. More precisely, our data generators are similar to those proposed by Wang \& al. \citep{wang2007group} and Barber \& al. \citep{barber2017function}. The basic concept is to build the functional response using a linear combination of a set of $K$ random curves $\psi_k(t)$ and $K$  associated  coefficients $\zeta_{k}(\boldsymbol{X})$: 

\begin{equation}
    Y(t) = \sum_{k=1}^K \zeta_{k}(\boldsymbol{X})\psi_k(t). 
    \label{gen_data}
\end{equation}
and we are going to explore multiple approaches to generate the random curves $\psi_k(t)$'s and multiple ways to use the predictors $\boldsymbol{X}$ to build coefficients $\zeta_{k}(\boldsymbol{X})$'s. 

Firstly, we concentrate on how to generate the curves $\psi_k(t)$'s. Wang \& al. \citep{wang2007group} employ spline functions constructed with B-splines. Due to the requirement of their simulation, they use rather simple order 4 B-splines with 1 interior knot corresponding to 5 basis functions. To create a simple scenario for visualization purpose, we consider a configuration where a random curve $\psi_{k}(t)$ is set to be a single B-spline basis function $B_{k}(t)$:
\begin{equation}
\psi_k(t) = B_{k}(t).
\end{equation}
Here, the applied B-spline basis system is of order of 4, while the number of interior knots are calculated using the number of predictors (number of predictors $-4$). The total number of basis functions, in this scenario, is equivalent to the number of scalar covariates. The main purpose of this configuration is to visualize if models are able to recover the coefficients $\zeta_{k}(\boldsymbol{X})$'s. Because in this example, a single covariate only affect a single B-spline basis function, we can fix $\zeta_{k}(\boldsymbol{X})$ to be a nonlinear function of the coefficients $\boldsymbol{X}$ and to see if the trained model captured that nonlinear effect. We notably use that configuration for our \textbf{Design 1}.

The second configuration is a somewhat more realistic case where the covariates affect the response over the entire time interval. In this configuration, each random curve $\psi_k(t)$ will be built using B-splines of order 4 with 9 interiors knots corresponding to 13 basis function $\{B_l(t)\}_{l=1}^{13}$. The 13 associated coefficients $\beta_l$'s are randomly generated from a normal distribution. Thus, for the B-splines configuration we have:
\begin{equation}
\psi_k(t)  = \sum_{l=1}^{13} \beta_{k,l} B_{l}(t).
\end{equation}
This configuration is used in our \textbf{Design 2, 3 \& 4}.

In terms of the coefficients $\zeta_{k}(\boldsymbol{X})$'s we will look at three configurations. Because NNs are known to capture nonlinear relationships, it is important to design nonlinear functions $\zeta_{k}(\cdot)$'s. First, for visualization, we apply a polynomial function to the continuous predictors. Specifically, a subset of continuous covariates, denoted by $\boldsymbol{X}_{\text{poly}}$, is randomly selected and each coviariate in $\boldsymbol{X}_{\text{poly}}$ is further transformed by a polynomial function with either second or third degree (half of the selected covariates are processed by quadratic functions and the rest by cubic functions), and accordingly $\zeta_{k}(\boldsymbol{X}) = \text{polynomial}(X_{k})$ for $ X_{k} \in \boldsymbol{X}_{\text{poly}}$, otherwise $\zeta_{k}(\boldsymbol{X}) = X_{k}$. This configuration is used in \textbf{Design 1 \& 2} with a visualization example displayed in  \textbf{Design 1}. Our second configuration is also nonlinear but this time it is a bit more complex.  We define a 3-hidden-layer NN with sigmoid (which is nonlinear) activation functions and random weights. For this configuration, $\zeta_{k}(\boldsymbol{X})$ is the $k$-th output of an NN taking $\boldsymbol{X}$ as its input. These coefficients are used for \textbf{Design 3}. Finally, we consider the case where $Y(t)$ is a linear combination of curves with $\boldsymbol{X}$ being the coefficients directly. In other words, the $k$-th coefficient is simply the $k$-th predictor: $\zeta_{k}(\boldsymbol{X}) = X_k$. This is a scenario where we expect the FoS to outperform the models we propose, but we want to visit this example regardless. We call this the linear configuration and use it for \textbf{Design 4}.

Finally, we add a random noise, which is normally distributed with zero mean and a variance of 2 to every data point $Y_i(t_{ij})$.

\subsection{Results}\label{results}

Four different simulation designs are considered to illustrate the advantages of the proposed methods. For each design, we generated data sets of 2000 observations and randomly sampled 1800 training observations and 200 testing observations. This random subsampling procedure was repeated 20 times. We opted to experiment with 20 scalar predictors, different random curve configurations and different relations between the coefficients and predictors, either linear or nonlinear, all with some random noise. To mimic the real-world scenario, the actual observations for $Y(t)$ were simulated discretely at 40 equally spaced time points $\{t_{j}\}_{j=1}^{40}$ that entirely cover $\mathcal{T}= [0,1]$. Similar to the real application, we compare the proposed NN-based methods, including NNBB, NNSS, NNBR and NNSR, to the novel FoS model (we exclude FAM model because its training cost is tremendously high and based on the results of the initial replicates, it performs poorly compared to the other models). The predictive accuracy was evaluated using the MSEP on the test set. We also report the $p$-value of the two-sided paired $t$-test of the MSEPs of each of our methods to that of the FoS. 

\textbf{Design 1 (nonlinear scenario):} We generate 20 random predictors ($K=20$), where $X_{k}$'s are i.i.d. uniform random variables from $[a, b]$ with $a \in \{-4,-3, -2, -1, 0\}, b \in \{3,4,5,6,7\}$ for all $k$. We apply the polynomial transformation to 50\% of the continuous variables by setting $\zeta_{k}(\boldsymbol{X}) = \text{polynomical}(X_{k})$ with degree of 2 and 3 for $k= 8, 10, 12, 13, 14$ and $k = 1, 3, 4, 7, 9$, respectively, and $\zeta_{k}(\boldsymbol{X}) = X_{k}$ otherwise. The random curves $\psi_{k}(t)$'s, in this case, are set to be the B-spline basis functions as $\psi_{k}(t) = B_{k}(t)$ for all $k$, and accordingly the functional response is simulated as $Y(t) = \sum_{k=1}^{20}\zeta_{k}(\boldsymbol{X})B_{k}(t)$. 

\begin{table}[ht]
\begin{minipage}{\textwidth}
\centering
\caption{Table of MSEPs of 20 random test sets for various models with data generated by \textbf{Design 1}.}
\label{tab:design1}
\begin{tabular}{@{} c c c c c c @{}}
\toprule
Methods & FoS  & NNBB & NNSS & NNBR & NNSR  \\
\midrule
Mean & 49.20 & 5.14 & 5.98 & 4.95 & 5.91\\ 
Std. Dev. & 1.64 &4.11 & 0.23 & 0.11 & 0.19 \\
$p$-value & - & $<$2.2e-16 & $<$2.2e-16 & $<$2.2e-16 & $<$2.2e-16\\
\botrule
\end{tabular}

\end{minipage}
\end{table}

Table \ref{tab:design1} summarizes the means and SDs of the MSEPs on the same testing observations achieved by different models in 20 replicates for \textbf{Design 1}. In the nonlinear scenario, we observe that the predictive performances of all the NN-based models surpass that of the FoS Model under 1\% significance level. Two proposed models using basis coefficients as the output of NN exhibit the best-level performances in predicting the response curve, with smaller means of MSEPs compared to those of the NN-based models outputting FPC scores. This is expected because the functional response in this design is generated as a directly linear combination of the polynomial-transformed predictors and B-spline basis functions, and given this special setting, we are able to visualize the relations as some describable patterns for some selected coefficient-predictor pairs. This design highlights a situation where using FoS would be extremely problematic and where any of our proposed NN models would provide a significant improvement. 

\begin{figure}[ht]
    \centering
    \includegraphics[width=0.9\columnwidth]{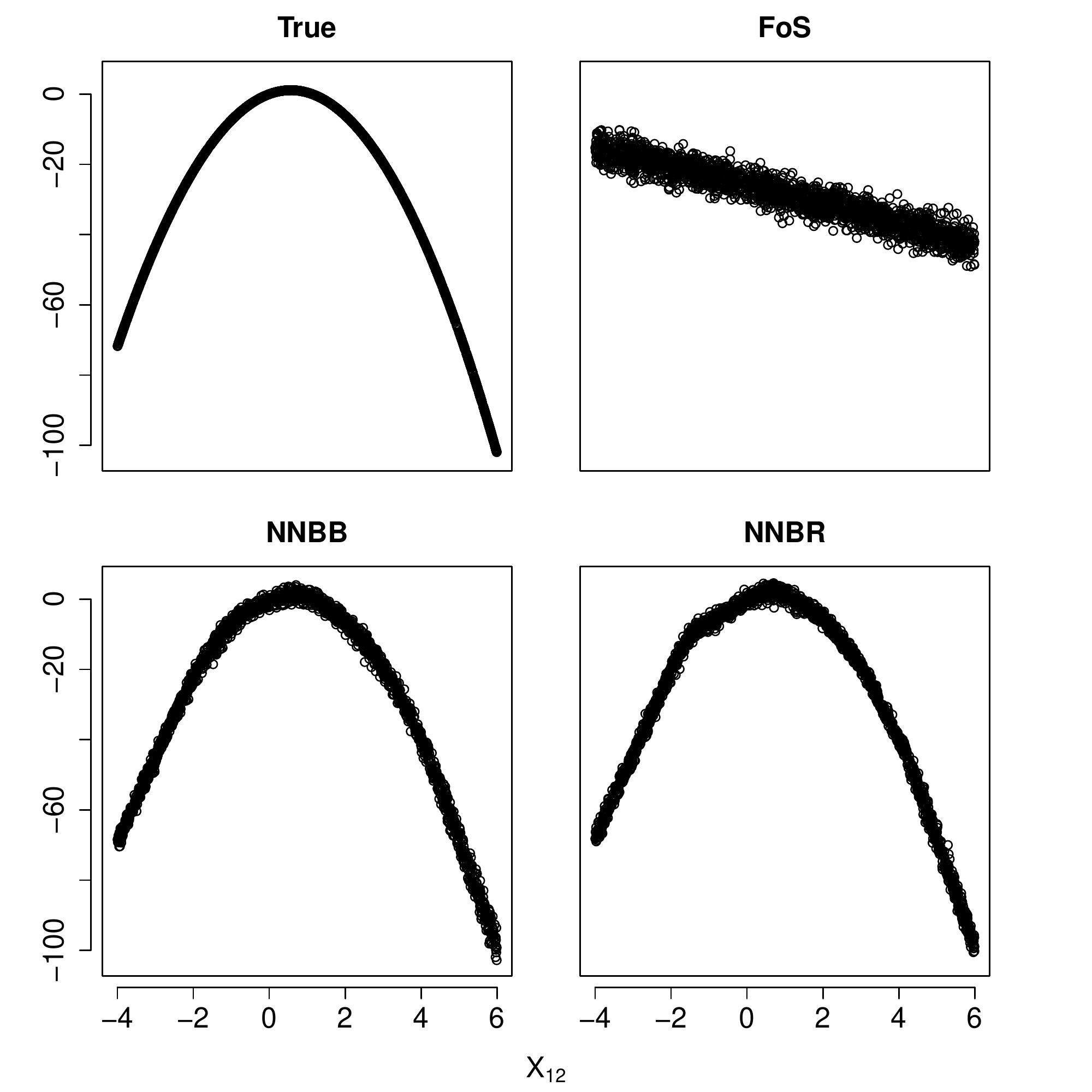}
    \caption{Scatter plots of the true $c_{12}$ ($\zeta_{12}$ as per the generator), FoS-predicted $\hat{c}_{12, \text{FoS}}$, NNBB-predicted $\hat{c}_{12, \text{NNBB}}$, and NNBR-predicted $\hat{c}_{12, \text{NNBR}}$ against  $X_{12}$ in \textbf{Design 1}, from left to right respectively.}
    \label{fig:design1_nonlinear}
\end{figure}

\begin{figure}[ht]
    \centering
    \includegraphics[width=0.9\columnwidth]{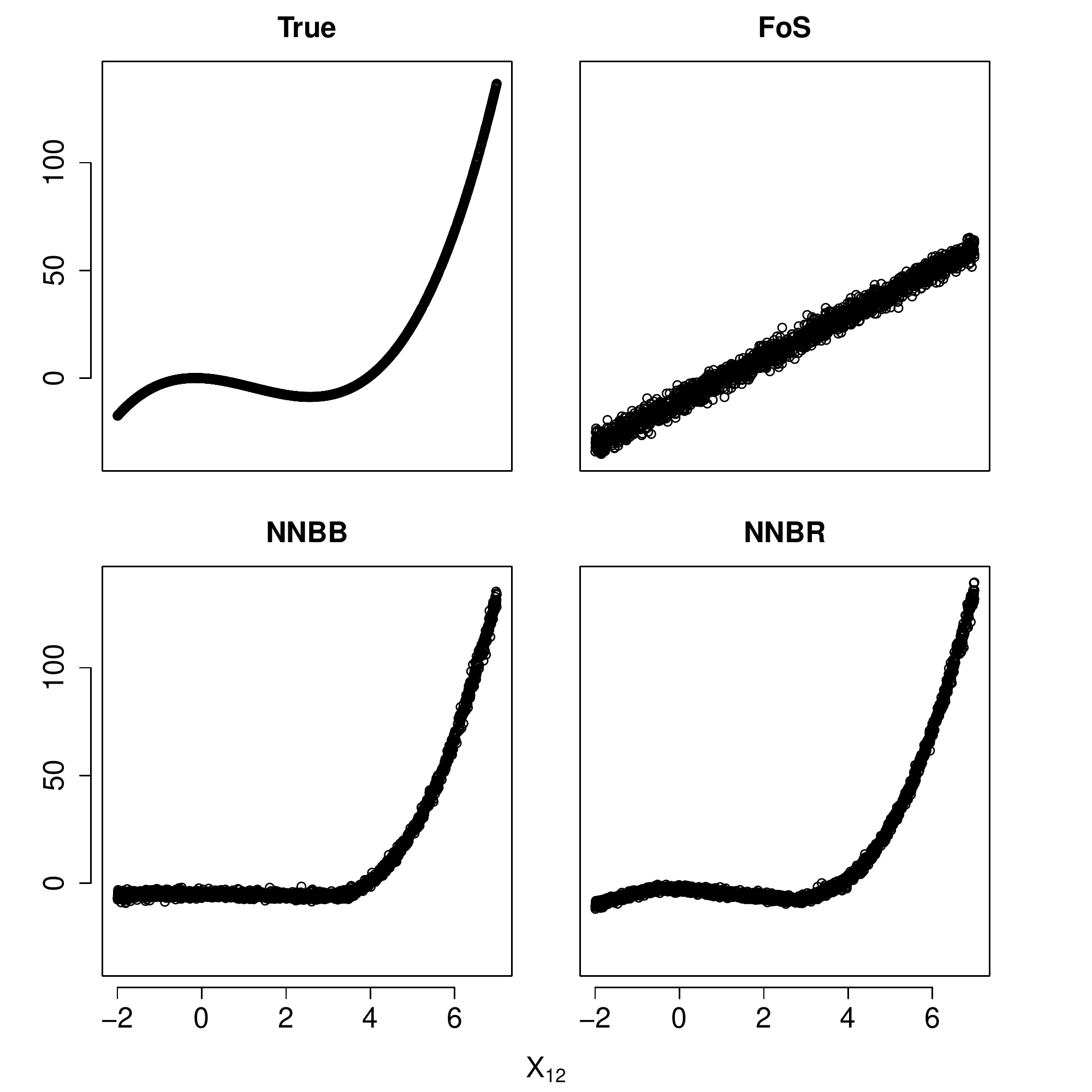}
    \caption{Scatter plots of the true $c_{4}$ ($\zeta_{4}$ as per the generator), FoS-predicted $\hat{c}_{4, \text{FoS}}$, NNBB-predicted $\hat{c}_{4, \text{NNBB}}$, and NNBR-predicted $\hat{c}_{4, \text{NNBR}}$ against  $X_{4}$ in \textbf{Design 1}, from left to right respectively.}
    \label{fig:design1_cubic}
\end{figure}

\begin{figure}[ht]
    \centering
    \includegraphics[width=0.9\columnwidth]{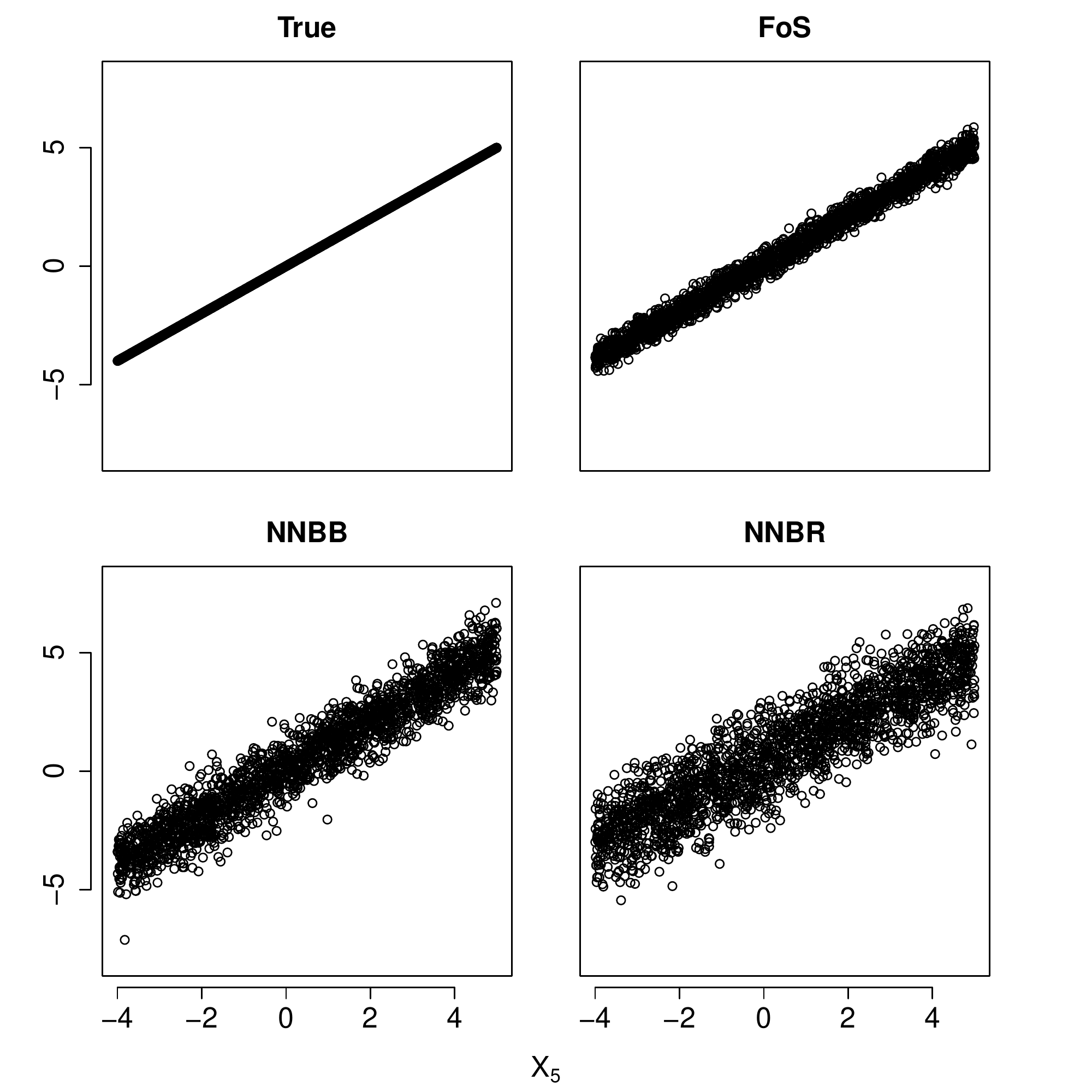} 
    \caption{Scatter plots of the true $c_{5}$ ($\zeta_{5}$ as per the generator), FoS-predicted $\hat{c}_{5, \text{FoS}}$, NNBB-predicted $\hat{c}_{5, \text{NNBB}}$, and NNBR-predicted $\hat{c}_{5, \text{NNBR}}$ against $X_{5}$ in \textbf{Design 1}, from left to right respectively.}
    \label{fig:design1_linear}
\end{figure}

Next, we compare NNBB, NNBR with FoS for their abilities
in recovering the underlying relationships for different coefficient-predictor pairs. The relations between a predictor and basis coefficients achieved by the aforementioned models are visualized through the scatter plots of each of the basis coefficients against the predictor. Remember that we designed this simulation specifically to visualize how NN models learn $\boldsymbol{C} = \text{NN}_{\eta}(\boldsymbol{X})$ and to compare it with the $\zeta_{k}(\boldsymbol{X})$ we designed.

We randomly select $X_{12}$, one of the polynomial-transformed covariates, and $c_{12}$, the basis coefficient corresponding to the $12$-th B-spline basis function, and show their true relation, as $\zeta_{12}(\cdot)$, together with the relations reconstructed by the three mentioned approaches using scatter plots of true $c_{12}$ and $\hat{c}_{12}$ predicted by each of the models against $X_{12}$ in Figure \ref{fig:design1_nonlinear}. Undoubtedly, NNBB and NNBR both precisely capture the $\cap$-shape between the true $c_{12}$ and $X_{12}$, while FoS replaces the nonlinear trend with a downwards linear pattern. Being interested in a more complex truth, we also select $X_4$, one of the covariates transformed with a cubic polynomial, coupled with $c_4$, the basis coefficient corresponding to the 4-th B-spline, and provide the scatter plots illustrating the true and model-reconstructed relations between $X_4$ and $c_4$ in Figure \ref{fig:design1_cubic} . As expected, NNBB and NNBR perform well in overall recovering the nonlinear shape while NNBR behaves more accurately in capturing the local curvature occurring in the left component of the true pattern. It is not surprising that FoS again learns a linear relation because it is designed to only learn such relations. Likewise, the associations between $X_{5}$ and $c_{5}$'s by multiple models are revealed in Figure \ref{fig:design1_linear}, but differently, the true relation for this $X_{5}$-$c_{5}$ pair is linear. We can observe, NNBB and NNBR perform similarly as FoS in retrieving the linear pattern but with comparably thicker bandwidths, indicating that our models can also successfully detect the linear relation for some coefficient-predictor pairs but with higher variance. Figure \ref{fig:design1_nonlinear}, \ref{fig:design1_cubic} and \ref{fig:design1_linear} highlight the utility of our methods in recovering the true underlying relations, especially the nonlinear relations between the basis coefficients and the predictors, and the success in learning the true relationships contributes to their fabulous performances on predicting the response trajectories. The ability to recover the true underlying nonlinear relations is the main benefit of our proposed approaches.

\textbf{Design 2 (nonlinear scenario):} 20 random predictors ($K=20$) are generated, with $X_{k}$'s being binary variables for $k=1, 3, 5, 7$, eight-level categorical variables for $k = 2, 4, 6$, and i.i.d. uniform random variables from $[a, b]$ with $a \in \{-4,-3, -2, -1, 0\}, b \in \{3,4,5,6,7\}$ for $k\geq 8$. Likewise, approximate 50\% of the continuous variables are later transformed by polynomial functions with different degrees, where $\zeta_{k}(X) = \text{polynomical}(X_{k})$ with the second and the third degree for $k=14, 16, 17, 18$ and $k=8, 10, 19, 20$, separately, and $\zeta_{k}(X) = X_{k}$ for the remainder. Then we construct the response curve as $Y(t) = \sum_{k=1}^{20}\zeta_{k}(\boldsymbol{X})\psi_{k}(t)$, coupled with $\psi_{k}(t) = \sum_{l=1}^{13}\beta_{k,l}B_{l}(t)$, where $\{B_{l}(t)\}_{l=1}^{13}$ are the B-spline basis functions with order 4 and $\beta_{k,l}$ are i.i.d. random variables following the normal distribution $\mathcal{N}(0, 4)$. 

\begin{table}[ht]
\begin{minipage}{\textwidth}
\centering
\caption{Table of MSEPs  of 20 random test sets for various models with data generated by \textbf{Design 2}.}
\label{tab:design2}
\begin{tabular}{@{} c c c c c c @{}}
\toprule
Methods & FoS  & NNBB & NNSS & NNBR & NNSR  \\
\midrule
Mean & 4559.20 & 38.33 & 286.23 & 36.38 & 265.50 \\ 
Std. Dev. & 159.01 & 9.18 & 79.68  & 18.76 & 16.05 \\
$p$-value & - & $<$2.2e-16 & $<$2.2e-16 & $<$2.2e-16 & $<$2.2e-16\\
\botrule
\end{tabular}
\end{minipage}
\end{table}

The means and SDs of the MSEPs on the testing observations for all methods in comparison can be found in Table \ref{tab:design2}. We observe that all our proposed methods remain superior to the FoS method in this setting. The NN-based models with basis-coefficient output continue to be the top-tier performers, especially the NNBR model trained using the response variable directly. The performance of the FoS is still significantly lower than the NN models and still has the highest variance. For both of these designs, some coefficient functions $\zeta_{k}$'s have significant nonlinear curvature being polynomial functions of degree 2 or 3. Consequently, the dominance of the NN models is expected, even now with a more realistic scenario where a predictor affects the response on its entire domain $\mathcal{T}$. 


\textbf{Design 3 (nonlinear scenario):} We continue to set $K=20$, and $X_{k}$'s are generated as binary variables, eight-level categorical variables and i.i.d. uniform random variables from $[a, b]$ with $a \in \{-4,-3, -2, -1, 0\}, b \in \{3,4,5,6,7\}$ for different sets of $k$ (same as \textbf{Design 2}). We trigger the nonlinear configuration by passing the predictors $\boldsymbol{X}$ through the 3-hidden-layer NN introduced in Section \ref{generator}, resulting in $\zeta({\boldsymbol{X}}) = \text{NN}(\boldsymbol{X})$. Same as \textbf{Design 2}, we choose to use the random curves $\psi_{k}(t) = \sum_{l=1}^{13}\beta_{k,l}B_{l}(t)$, where $\{B_{l}(t)\}_{l=1}^{13}$ are the B-spline basis functions with order 4 and $\beta_{k,l} \overset{\text{i.i.d.}}{\sim} \mathcal{N}(0, 4)$. In consequence, the functional response is generated following $Y(t) = \sum_{k=1}^{20}\text{NN}(\boldsymbol{X})\psi_{k}(t)$.

\begin{table}[ht]
\begin{minipage}{\textwidth}
\centering
\caption{Table of MSEPs  of 20 random test sets for various models with data generated by \textbf{Design 3}.}
\label{tab:design3}
\begin{tabular}{@{} c c c c c c @{}}
\toprule 
Methods & FoS  &  NNBB & NNSS & NNBR & NNSR  \\
\midrule
Mean & 4.17 & 4.18 &  4.16 & 4.13 &  4.17  \\
Std. Dev. & 0.02 &  0.03  & 0.02 & 0.03  & 0.02 \\
$p$-value & - & 1.1e-02 & 6.2e-02& 2.7e-08 & 1.8e-01 \\
\botrule
\end{tabular}
\end{minipage}
\end{table}

Table \ref{tab:design3} reports the means and SDs of MSEPs on the test sets for all models with data generated by \textbf{Design 3}. It is shown that most models perform similarly on predicting the functional curve. The $\zeta_k(\boldsymbol{X})$'s produced by the NN generator are barely nonlinear, having only small curvature with respect to the predictors. This leads to a reduction in the performance gap between our NN models and the FoS model. However, the NNBR model comes ahead once again, being significantly superior to the FoS model. 


\textbf{Design 4 (linear scenario):} In the last design, we consider setting up a scenario where the functional response $Y(t)$ is linearly associated to the scalar predictors. 20 predictors are generated in the manner that $\{X_{k}\}_{k=1}^{20}$ are binary variables, eight-level categorical variables and i.i.d. uniform random variables from $[a, b]$ with $a \in \{-4,-3, -2, -1, 0\}, b \in \{3,4,5,6,7\}$ for different subsets of $k = {1, 2, ..., 20}$ (same as \textbf{Design 2 \& 3}). The linear setting is simply achieved with $\zeta_{k}({\mathbf{X}}) = X_{k}$. We continue with the random curves $\psi_{k}(t) = \sum_{l=1}^{13}\beta_{k,l}B_{l}(t)$, where $\{B_{l}(t)\}_{l=1}^{13}$ are the order 4 B-spline basis functions, together with $\beta_{k,l} \overset{\text{i.i.d.}}{\sim} \mathcal{N}(0, 4)$. The functional response is generated as a linear combination of the random curves and predictors, following $Y(t) = \sum_{k=1}^{20}X_{k}\psi_{k}(t)$.

\begin{table}[ht]
\begin{minipage}{\textwidth}    
\centering
\caption{Table of MSEPs  of 20 random test sets for various models with data generated by \textbf{Design 4}.}
\label{tab:design4}
\begin{tabular}{@{} c c c c c c @{}}
\toprule
Methods & FoS  &  NNBB & NNSS & NNBR & NNSR  \\
\midrule 
Mean & 4.01  & 4.19 & 6.66 & 4.07 & 5.94 \\
Std. Dev. & 0.05  & 0.06 & 0.15 & 0.05 & 0.05 \\
$p$-value & - & 3.2e-08 & 4.3e-13 & 7.8e-07 & 2.2e-09 \\
\botrule
\end{tabular}
\end{minipage}
\end{table}

Table \ref{tab:design4}
displays the results of all models under the linear setting described. In the linear scenario, we expect FoS to be the top performer and indeed, it has the lowest prediction error compared to all the proposed methods. This is reasonable as the $Y(t)$ is constructed linearly with respect to the predictors in this design. However, despite the proposed approaches being significantly worse, NNBB and NNBR remain competitive with strong performances closely following the one of FoS. From an absolute perspective, the gap between NNBR and FoS is much smaller in the scenario favouring FoS (\textbf{Design 4}), with FoS being marginally ahead, than it is in the scenario favouring NNBR (\textbf{Design 1 \& 2}), with NNBR being dramatically ahead. 




\section{Conclusions and Discussion}\label{sec:conclusion}

In the article, we introduced a new solution for the regression of a functional response on scalar predictors, which consistently outperforms the current models when the relation between the functional response and the scale predictors is nonlinear. We designed and manipulated the standard feed-forward NN to produce two types of output, either basis coefficients or FPC scores, both of them being the traditional techniques to analyse functional data. The proposed modification to the objective function allows us to train the model directly with the functional response variable thus bypassing the necessity to firstly estimate those coefficients using conventional FDA approaches. The modified objective function enables the transformation of a scalar layer to a functional output by constructing functional curves using a linear operation, which ensures the applicability of back-propagation. In this way, an NN with the proposed functional output layer behaves similarly as a regular NN and allows the usage of typical cross-validation techniques for hyper-parameter tuning. Such modifications can be directly implemented to the output layer of any existing NN to produce a functional response, promoting the utilization of various deep learning techniques in a functional regression framework. We implemented all of our proposed models in a way they can be trained on both regularly and irregularly spaced domain. Additionally, we provided the necessary tools to control the smoothness of the predicted response curves by implementing two different roughness penalties. Furthermore, our family of models scales better with the number of predictors. 

Based on a real data application, we demonstrated a case where the models we propose are superior to already established techniques such as FoS and FAM. Moreoever, through several simulation studies, we not only showed the superior predictive power of our NN-based approaches, but also their strong ability to recover nonlinear relation between the predictors and the coefficients representing the response curves.

On the other hand, the developed methods rely on a large number of hyper-parameters, including the number of hidden layers, the number of neurons in each of the hidden layers, the number of basis functions (NN output size), the number of training epochs, etc. This is the main weakness of our approach since conducting a grid search on that space is particularly annoying and time-consuming. For both B-spline expansion models, the performance varied dramatically from one hyper-parameter configuration to another and this is certainly what we need to bring up. Comparatively, the FPCA-based models varied only slightly while changing the number of principal components. Given once we have the first few leading principal components, we can capture the vast majority of the variability between curves. The performances were also much more stable across various parameterizations, though those models were rarely top performers. In contrast, the FoS model is the easiest to fit among all the models applied, albeit it performed poorly in nonlinear situations. 

Multiple directions of further research are considered at the moment. Our proposed models can be extended to predict multidimensional (mainly two-dimensional) functional response where $t \in \mathbb{R}^{q}, q>1$. In this scenario, we can borrow the basis expansion or the FPCA technique to compress the multidimensional functional data to a vector of finite scalar basis coefficients or FPC scores \citep{smoothing_2Dfd, surface_fitting, surface_fitting2, FPCA_2Dfd, FPCA_MDfd}. Additionally, our current models rely on the existing NN architecture with a scalar output layer, which requires us to firstly project the functional response to some finite-dimensional coefficients and then feed the NN with the scalar representations obtained. Extending the NN to a more dynamic architecture allowing a functional output layer could be a more appropriate tool for such type of regression problems. This could be achieved by defining functional neurons and functional layers, some concepts we are currently exploring. Furthermore, a combination of our proposed NN models with existing literature that tackles the SoF problem can be explored to create a general and complete framework for using NN to analyze and solve regression problems for various forms of functional data. 

\section*{Supplementary Materials}
The supplementary document includes a list of all notations used in the manuscript and model configurations used in the application.
We also provide the \texttt{R} codes and data required
to reproduce the numerical results in the real application.

\section*{Acknowledgements}

The authors would like to acknowledge the financial support of the Canadian Statistical Sciences Institute (CANSSI) and the Natural Sciences and Engineering Research Council of Canada (NSERC). \\

\noindent \textbf{Declarations}: The authors have no conflicts of interest to declare.  

\bibliographystyle{plain}
\bibliography{sn-bibliography}


\end{document}